\documentclass[10pt,twocolumn,letterpaper]{article}

 \usepackage{cvpr}              % To produce the CAMERA-READY version
\usepackage[accsupp]{axessibility}
\usepackage{amssymb}
\usepackage{amsmath}
\usepackage{graphicx}
\usepackage{algorithm}
\usepackage{algpseudocode}
\usepackage{multirow}
\newtheorem{theorem}{\textbf{Theorem}}
\newtheorem{definition}{\textbf{Definition}}
\newtheorem{remark}{\textbf{Remark}}
\newtheorem{property}{\textbf{Property}}
\newcommand{\argmin}{\mathop{\mathrm{argmin}}}

\usepackage{tabularx}

\usepackage[dvipsnames]{xcolor}

\definecolor{cvprblue}{rgb}{0.21,0.49,0.74}
\usepackage[pagebackref,breaklinks,colorlinks,citecolor=cvprblue]{hyperref}

\def\paperID{13742} % *** Enter the Paper ID here
\def\confName{CVPR}
\def\confYear{2024}

\title{SVDinsTN: A Tensor Network Paradigm for Efficient Structure Search\\ from Regularized Modeling Perspective}

\author{Yu-Bang Zheng$^{1}$\ \ \ \ \ \ Xi-Le Zhao$^{2,}$\thanks{Corresponding author.}\ \ \ \ \ \ \ Junhua Zeng$^{3,4}$\ \ \ \ \ \ Chao Li$^{4}$\\
	Qibin Zhao$^{4}$\ \ \ \ \ \ Heng-Chao Li$^{1}$\ \ \ \ \ \ Ting-Zhu Huang$^{2}$\\
$^{1}$School of Information Science and Technology, Southwest Jiaotong University, China\\
$^{2}$School of Mathematical Sciences, University of Electronic Science and Technology of China, China\\	
$^{3}$School of Automation, Guangdong University of Technology, China\\
$^{4}$Tensor Learning Team, RIKEN Center for Advanced Intelligence Project (AIP), Japan\\
{\tt\small zhengyubang@163.com, xlzhao122003@163.com, jh.zenggdut@gmail.com, chao.li@riken.jp}\\
{\tt\small qibin.zhao@riken.jp, hcli@home.swjtu.edu.cn, tingzhuhuang@126.com}
}

\begin{document}
\maketitle

\begin{abstract}
	Tensor network (TN) representation is a powerful technique for computer vision and machine learning. TN structure search (TN-SS) aims to search for a customized structure to achieve a compact representation, which is a challenging NP-hard problem. Recent ``sampling-evaluation''-based methods require sampling an extensive collection of structures and evaluating them one by one, resulting in prohibitively high computational costs. To address this issue, we propose a novel TN paradigm, named SVD-inspired TN decomposition (SVDinsTN), which allows us to efficiently solve the TN-SS problem from a regularized modeling perspective, eliminating the repeated structure evaluations. To be specific, by inserting a diagonal factor for each edge of the fully-connected TN, SVDinsTN allows us to calculate TN cores and diagonal factors simultaneously, with the factor sparsity revealing a compact TN structure. In theory, we prove a convergence guarantee for the proposed method. Experimental results demonstrate that the proposed method achieves approximately $100\!\sim{}\!1000$ times acceleration compared to the state-of-the-art TN-SS methods while maintaining a comparable level of representation ability.  
\end{abstract}

\section{Introduction}

Tensor network (TN) representation, which aims to express higher-order data with small-sized tensors (called TN cores) by a specific operation among them, has gained significant attention in various areas of data analysis~\cite{Anandkumar,Han_2023_CVPR,RaiRNK,BayesianCP}, machine learning~\cite{MAL_059,NEURIPGlasser,Shakeri_2019_CVPR}, computer vision~\cite{9880157,9878497,9577826,8578957,ZHENG2023182}, etc. By regarding TN cores as nodes and operations as edges, a TN corresponds to a graph (called TN topology). The vector composed of the weights of all edges in the topology is defined as the TN rank. \textit{TN structure (including topology and rank) search (TN-SS)} aims to search for a suitable TN structure to achieve a compact representation for a given tensor, which is known as a challenging NP-hard problem~\cite{MostNP,li2023alternating}. The selection of TN structure dramatically impacts the performance of TN representation in practical applications~\cite{LiTopologySearch,hashemizadeh2022adaptive,LiPermutationSearch,li2023alternating}. 

Recently, there have been several notable efforts to tackle the TN-SS problem~\cite{hashemizadeh2022adaptive,LiTopologySearch,LiPermutationSearch,li2023alternating}. But most of them adopt the ``sampling-evaluation'' framework, which requires sampling a large number of structures as candidates and conducting numerous repeated structure evaluations. For instance, for a tensor of size $40\times60\times3\times9\times9$ (used in Section~\ref{seccr}), TNGA in~\cite{LiTopologySearch} requires \textit{thousands} of evaluations and TNALE in~\cite{li2023alternating} requires \textit{hundreds} of evaluations, where each evaluation entails solving an optimization problem to compute TN cores iteratively. Consequently, the computational cost becomes exceedingly high. A meaningful question is \textit{whether we can optimize the TN structure simultaneously during the computation of TN cores, thus escaping the ``sampling-evaluation'' framework and fundamentally addressing the computationally consuming issue.}

\begin{figure*}
	\centering
	\includegraphics[width=0.92\textwidth]{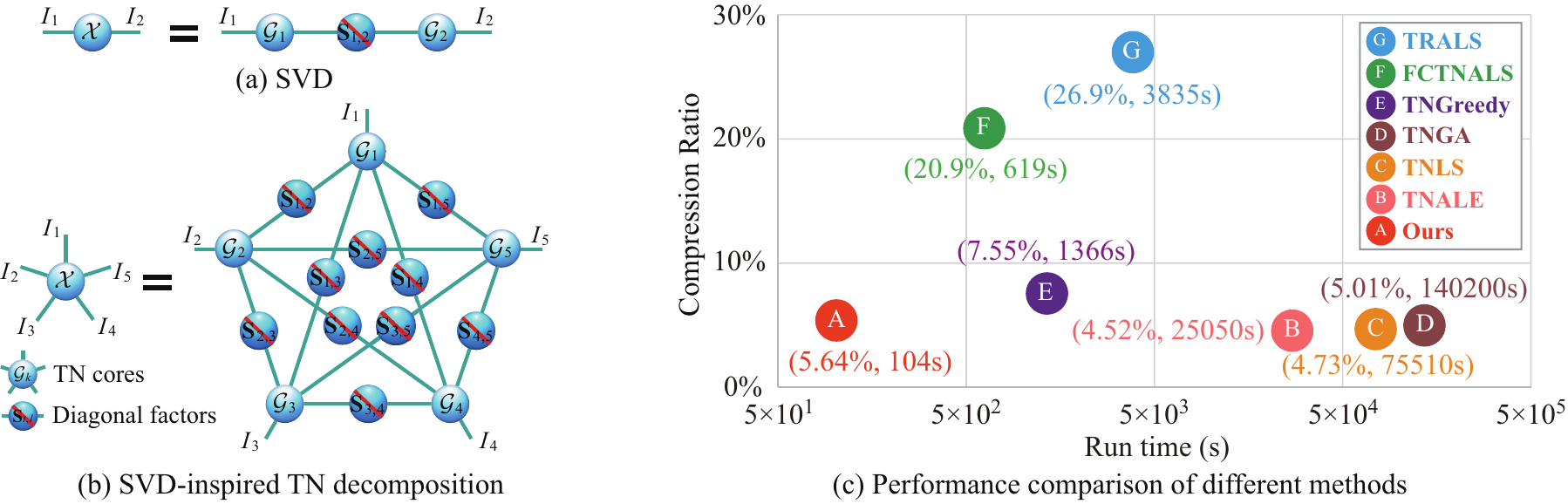}
	\caption{(a) A graphical illustration of SVD. (b) A graphical illustration of SVD-inspired TN decomposition on a fifth-order tensor. (c) Comparison of the compression ratio ($\downarrow$) and run time ($\downarrow$) of different methods on a fifth-order light field image \textit{Knights}, where the reconstruction error bound is set to 0.05, TRALS~\cite{zhao2016tensor} and FCTNALS~\cite{zhengFCTN} are methods with pre-defined topologies, and TNGreedy~\cite{hashemizadeh2022adaptive}, TNGA~\cite{LiTopologySearch}, TNLS~\cite{LiPermutationSearch}, and TNALE~\cite{li2023alternating} are TN-SS methods (please see more results in Table~\ref{FLpertab}).}
	\label{illu_ATN}
\end{figure*}

In this paper, we introduce for the first time a regularized modeling perspective on solving the TN-SS problem. This perspective enables us to optimize the TN structure simultaneously during the computation of TN cores, effectively eliminating the need for repetitive structure evaluations. To be specific, we propose a novel TN paradigm, termed as \textit{SVD-inspired TN decomposition (SVDinsTN)}, by inserting diagonal factors between any two TN cores in the ``fully-connected'' topology (see Figure~\ref{illu_ATN}(b)). \textit{The intuition behind SVDinsTN is to leverage the sparsity of the inserted diagonal factors to reveal a compact TN structure and utilize the TN cores (merged with the diagonal factors) to represent a given tensor.} Based on SVDinsTN, we establish a regularized model, which updates the TN cores and diagonal factors iteratively and imposes a sparse operator to induce the sparsity of the diagonal factors. In theory, we prove a convergence guarantee for the proposed method and establish an upper bound for the TN rank. In particular, we design a novel initialization scheme for the proposed method based on the upper bound. This initialization scheme enables the proposed method to overcome the high computational cost in the first several iterations, which is caused by the utilization of a ``fully-connected'' topology as the starting point. As a result, SVDinsTN is capable of capturing a customized TN structure and providing a compact representation for a given tensor in an efficient manner. In summary, we make the following three contributions.

\begin{itemize}
	
	\item We propose SVDinsTN, a new TN paradigm, that enables us to optimize the TN structure during the computation of TN cores, greatly reducing the computational cost. 
	
	\item In theory, we prove a convergence guarantee for the proposed method and establish an upper bound for the TN rank involved in SVDinsTN. The upper bound serves as a guide for designing an efficient initialization scheme.
	
	\item Experimental results verify numerically that the proposed method achieves $100\!\sim\!1000$ times acceleration compared to the state-of-the-art TN-SS methods with a comparable representation ability (see Figure~\ref{illu_ATN}(c)).
\end{itemize}

\subsection{Related Works}

\textbf{TN representation}\footnote{We focus on TN representation in scientific computing and machine learning, while acknowledging its history of research in physics~\cite{PhysRevATT,ORUS2014117,YeKE}.} aims to find a set of small-sized TN cores to express a large-sized tensor under a given TN structure (including topology and rank)~\cite{CichockiSPM,MAL_059,YeKE}. In the past decades, many works focused on TN representation with a fixed TN topology, such as tensor train (TT) decomposition with a ``chain'' topology~\cite{SIAMTT}, tensor ring (TR) decomposition with a ``ring'' topology~\cite{zhao2016tensor}, fully-connected tensor network (FCTN) decomposition with a ``fully-connected'' topology~\cite{zhengFCTN}, etc. In addition, these works also presented various methods to optimize the TN cores, such as alternating least square (ALS)~\cite{zhao2016tensor}, gradient descent (GD)~\cite{Yuan2018,SGDTTYuan}, proximal alternating minimization (PAM)~\cite{zhengFCTN,zheng2022tensor}, etc. In contrast, SVDinsTN can reveal a compact TN structure for a given tensor, surpassing methods with pre-defined topologies in terms of representation ability.

\begin{table*}[!t]
	\small
	\renewcommand\arraystretch{1}
	\caption{Several operations and their interpretations.}\label{interpretations}
	\centering
	\begin{tabularx}{\textwidth}{lX}
		\toprule
		Operation & Interpretation\\
		
		\midrule
		$\mathrm{diag}$  & $\mathrm{diag}(\mathbf{X})$ returns a column vector formed from the elements on the main diagonal of $\mathbf{X}$ when the input variable is a diagonal matrix, and $\mathrm{diag}(\mathbf{x})$ returns a diagonal matrix whose main diagonal is formed from the elements of $\mathbf{x}$ when the input variable is a column vector. \\
		$\mathrm{ones}$& $\mathrm{ones}(I_1,I_2,\cdots,I_N)$ returns an $I_1\times I_2\times\cdots\times I_N$ tensor, whose elements are all equal to 1.\\
		$\mathrm{zeros}$& $\mathrm{zeros}(I_1,I_2,\cdots,I_N)$ returns an $I_1\times I_2\times\cdots\times I_N$ tensor, whose elements are all equal to 0.\\
		$\mathrm{vec}$& $\mathrm{vec}(\mathcal{X})$ returns a column vector by lexicographical reordering of the elements of $\mathcal{X}$.\\		
		\bottomrule
	\end{tabularx}
\end{table*}

\textbf{TN structure search (TN-SS)} aims to search for a suitable or optimal TN structure, including both topology and rank, to achieve a compact representation for a given tensor~\cite{SedighinTRrank,Nie2021AdaptiveTN,hashemizadeh2022adaptive,LiTopologySearch,LiPermutationSearch,HeuristicrankTR,li2023alternating,ghadiri2023approximately,liu2023adaptively}. However, the majority of existing TN-SS methods follow the ``sampling-evaluation'' framework, which necessitates the use of heuristic search algorithms like the greedy algorithm~\cite{hashemizadeh2022adaptive}, genetic algorithm~\cite{LiTopologySearch}, and alternating local enumeration algorithm~\cite{li2023alternating} to sample candidate structures and subsequently evaluate them individually. Therefore, these methods inevitably suffer from prohibitively high computational costs due to the numerous repeated evaluations, each involving the iterative calculation of TN cores within an optimization problem.
In contrast, SVDinsTN addresses the TN-SS problem from a regularized modeling perspective, thereby avoiding the repeated structure evaluations and significantly reducing computational costs.

\section{Notations and Preliminaries}\label{sec_nota}

A \textit{tensor} is a multi-dimensional array, and the number of dimensions (also called \textit{modes}) of which is referred to as the \textit{tensor order}. 
In the paper, first-order tensors (vectors), second-order tensors (matrices), and $N$th-order tensors are denoted by $\mathbf{x}\in \mathbb{R}^{I_1}$, $\mathbf{X}\in \mathbb{R}^{I_1\times I_2}$, and $\mathcal{X}\in \mathbb{R}^{I_1\times I_2\times\cdots\times I_N}$, respectively. We use $\|\mathcal{X}\|_F$ and $\|\mathcal{X}\|_1$ to denote the Frobenius norm and $\ell_1$-norm of $\mathcal{X}$, respectively. To simplify the explanation, we let $x_{1:d}$ denote the ordered set $\{x_{1},x_{2},\cdots,x_{d}\}$,
$\mathbb{K}_{N}$ denote the set $\{1,2,\cdots,N\}$, and $\mathbb{TL}_{N}$ denote the set $\{(t,l)|1\leq t<l\leq N;t,l\in \mathbb{N}\}$.

We next review several operations on tensors~\cite{zhengFCTN}.

The \textit{generalized tensor transposition}~\cite{zhengFCTN} is an operation that rearranges tensor modes. For example, an $I_1\times I_2\times I_3\times I_4$ tensor can be transposed into an $I_3\times I_2\times I_1\times I_4$ tensor, denoted by $\mathcal{\vec{X}}^{\mathbf{n}}$ with $\mathbf{n}=(3,2,1,4)$.  We use $\mathcal{\vec{X}}^{\mathbf{n}}=\mathrm{permute}(\mathcal{X},\mathbf{n})$ and $\mathcal{X}=\mathrm{ipermute}(\mathcal{\vec{X}}^{\mathbf{n}},\mathbf{n})$ to denote the corresponding transposition operation and its inverse operation, respectively.

The \textit{generalized tensor unfolding}~\cite{zhengFCTN} is an operation that converts a tensor into a matrix by merging a group of tensor modes into the rows of the matrix and merging the remaining modes into the columns. For example, an $I_1\times I_2\times I_3\times I_4$ tensor can be unfolded into an $I_1I_3\times I_2I_4$ matrix. We use $\mathbf{X}_{[1,3;2,4]}=\mathrm{GUnfold}(\mathcal{X}, (1,3;2,4))$ and $\mathcal{X}=\mathrm{GFold}(\mathbf{X}_{[1,3;2,4]}, (1,3;2,4))$ to denote the corresponding unfolding operation and its inverse operation, respectively. We also use $\mathbf{X}_{(2)}$ to simply denote $\mathbf{X}_{[2;1,3,4]}\in\mathbb{R}^{I_2\times I_1I_2I_4}$, which is also called \textit{mode-2 unfolding}. We use $\mathbf{X}_{(2)}=\mathrm{Unfold}(\mathcal{X},2)$ and $\mathcal{X}=\mathrm{Fold}(\mathbf{X}_{(2)},2)$ to denote the corresponding mode-2 unfolding operation and its inverse operation, respectively~\cite{kolda2009tensor}. 

The \textit{tensor contraction}~\cite{zhengFCTN} is an operation that obtains a new tensor by pairing, multiplying, and summing indices of certain modes of two tensors. For example, if a fourth-order tensor $\mathcal{X}\in \mathbb{R}^{I_1\times I_2\times I_3\times I_4}$ and a third-order tensor $\mathcal{Y}\in \mathbb{R}^{J_1\times J_2\times J_3}$ satisfy $I_{2}=J_{1}$ and $I_{4}=J_{2}$, then the tensor contraction between the 2nd and 4th modes of $\mathcal{X}$ and the 1st and 2nd modes of $\mathcal{Y}$ yields a tensor $\mathcal{Z}=\mathcal{X}\times_{2,4}^{1,2}\mathcal{Y}\in\mathbb{R}^{I_1\times I_3\times J_3}$. The elements of $\mathcal{Z}$ are calculated as follows:
\begin{equation*}
	\mathcal{Z}(i_{1},i_{3}, j_{3})\!=\!\sum\nolimits_{i_{2}=1}^{I_{2}}\!\sum\nolimits_{i_{4}=1}^{I_{4}}
	\!\mathcal{X}(i_{1},i_{2},i_{3},i_{4})
	\mathcal{Y}(i_{2},i_{4},j_{3}).
\end{equation*}

In Table~\ref{interpretations}, we give the interpretations of the operations  ``$\mathrm{diag}$'', ``$\mathrm{ones}$'', ``$\mathrm{zeros}$'', and ``$\mathrm{vec}$''.
\subsection{Tensor Network}

In general, a \textit{tensor network (TN)} is defined as a set of small-sized tensors, known as TN cores, in which some or all modes are contracted according to specific operations~\cite{MAL_059}. The primary purpose of a TN is to represent higher-order data using these TN cores. By considering TN cores as nodes and operations between cores as edges, we define \textit{the graph formed by these nodes and edges as the TN topology}. Additionally, we assign a non-negative integer weight to each edge to indicate the size of the corresponding mode of TN cores, and call \textit{the vector composed of these weights the TN rank}. Consequently, \textit{a TN structure refers to a weighted graph, encompassing both the TN topology and TN rank.}

This paper focuses on only a class of TNs that employs tensor contraction as the operation among TN cores and adopts a simple graph as the TN topology. More particularly, when representing an $N$th-order tensor $\mathcal{X}$, this class of TNs comprises precisely $N$ TN cores, each corresponding to one mode of $\mathcal{X}$. A notable method is FCTN decomposition, which represents an $N$th-order tensor $\mathcal{X}\in \mathbb{R}^{I_1\times I_2\times\cdots\times I_N}$ by $N$ small-sized $N$th-order cores denoted by $\mathcal{G}_k\in \mathbb{R}^{R_{1,k}\times R_{2,k}\times \cdots\times R_{k-1,k}\times I_k\times R_{k,k+1}\times\cdots\times R_{k,N}}$ for $k\in$ $ \mathbb{K}_{N}$~\cite{zhengFCTN}. In this decomposition, any two cores $\mathcal{G}_l$ and $\mathcal{G}_t$ for $(t,l)\in \mathbb{TL}_{N}$ share an equal-sized mode $R_{t,l}$ used for tensor contraction. We denote the above FCTN decomposition by $\mathcal{X}=\text{FCTN}(\mathcal{G}_{1:N})$ and define the FCTN rank as the vector $(R_{1,2},R_{1,3},\cdots, R_{1,N}, R_{2,3},\cdots,  R_{2,N},$ $\cdots, R_{N-1,N})\in \mathbb{R}^{{N(N-1)}/{2}}$. According to the concept of tensor contraction, removing rank-one edges in the TN topology does not change the expression of the TN. This means that if any element in the FCTN rank is equal to one, the corresponding edge can be harmlessly eliminated from the ``fully-connected'' topology. For instance, a ``fully-connected'' topology with the rank $(R_{1,2},1, \cdots, 1, R_{2,3},$ $1,\cdots,1,R_{N-2,N-1}, R_{N-1,N})$ can be converted into a ``chain'' topology with rank $(R_{1,2}, R_{2,3},\cdots,R_{N-1,N})$ in this manner. 
This fact can be formally stated as follows.

\begin{property}{\cite{Liworkshop}}\label{SandR}
	There exists a one-to-one correspondence between the \textit{TN structure} and \textit{FCTN rank}. 
\end{property}

According to Property~\ref{SandR}, we can search for a compact TN structure by optimizing the FCTN rank.

\section{An Efficient Method for TN-SS}  \label{secTNnet}

We propose an efficient method to solve the TN-SS problem from a regularized modeling perspective. Unlike the existing ``sampling-evaluation'' framework, the \emph{main idea} of the proposed method is to optimize the TN structure (the FCTN rank) simultaneously during the computation of TN cores, thereby eliminating the repetitive structure evaluations and greatly decreasing the computational cost.

\subsection{SVDinsTN}

We start with the definition of the following SVDinsTN.

\begin{definition}[SVDinsTN]\label{SFCTN} Let $\mathcal{X}\in\mathbb{R}^{I_1\times I_2\times\cdots\times I_N}$ be an $N$th-order tensor such that
	\begin{equation}
		\begin{aligned}
			&\mathcal{X}(i_1,i_2,\cdots,i_N)=\\
			&\sum_{r_{1,2}=1}^{R_{1,2}}\sum_{r_{1,3}=1}^{R_{1,3}}\cdots\sum_{r_{1,N}=1}^{R_{1,N}}
			\sum_{r_{2,3}=1}^{R_{2,3}}\cdots\sum_{r_{2,N}=1}^{R_{2,N}}\cdots\sum_{r_{N-1,N}=1}^{R_{N-1,N}}\\
			&\mathbf{S}_{1,2}(r_{1,2},r_{1,2})\mathbf{S}_{1,3}(r_{1,3},r_{1,3})\cdots\mathbf{S}_{1,N}(r_{1,N},r_{1,N})\\
			&\mathbf{S}_{2,3}(r_{2,3},r_{2,3})\cdots\mathbf{S}_{2,N}(r_{2,N},r_{2,N})\cdots\\
			&\mathbf{S}_{N-1,N}(r_{N-1,N},r_{N-1,N})\\
			&\mathcal{G}_1(i_1,r_{1,2},r_{1,3},\cdots,r_{1,N})\\
			&\mathcal{G}_2(r_{1,2},i_2,r_{2,3},\cdots,r_{2,N})\cdots\\
			&\mathcal{G}_k(r_{1,k},r_{2,k},\cdots,r_{k-1,k},i_k,r_{k,k+1},\cdots,r_{k,N})\cdots\\
			&\mathcal{G}_N(r_{1,N},r_{2,N},\cdots,r_{N-1,N},i_N),
		\end{aligned} \label{TNtensor3}
	\end{equation}	
	where $\mathcal{G}_k\in \mathbb{R}^{R_{1,k}\times R_{2,k}\times \cdots\times R_{k-1,k}\times I_k\times R_{k,k+1}\times\cdots\times R_{k,N}}$ for $\forall k\in \mathbb{K}_N$ are $N$th-order tensors and called TN cores, and $\mathbf{S}_{t,l}\in\mathbb{R}^{R_{t,l}\times R_{t,l}}$ for $\forall (t,l)\in\mathbb{TL}_{N}$ are diagonal matrices. Then we call (\ref{TNtensor3}) an SVD-inspired TN decomposition (SVDinsTN) of $\mathcal{X}$, denoted by $\mathcal{X}= \text{STN}(\mathcal{G},\mathbf{S})$, where $\mathcal{G}$ denotes $\{\mathcal{G}_{k}|k\in\mathbb{K}_N\}$ and $\mathbf{S}$ denotes $\{\mathbf{S}_{t,l}|(t,l)\in\mathbb{TL}_{N}\}$.
\end{definition}

As shown in Figure~\ref{illu_ATN}(b), SVDinsTN includes both TN cores and diagonal factors, and can use the sparsity of diagonal factors to reveal a compact TN structure and utilize TN cores (merged with diagonal factors) to represent a tensor.

\begin{remark}[SVDinsTN~\&~SVD] As shown in Figure~\ref{illu_ATN}(a)-(b), SVDinsTN extends the ``core~\&~diagonal factor~\&~core'' form of SVD to higher-order cases,  incorporating the idea of determining rank through non-zero elements in the diagonal factor. In particular, SVDinsTN can degrade into SVD in second-order cases when TN cores satisfy orthogonality.
\end{remark}

\begin{remark}[SVDinsTN \& FCTN] SVDinsTN builds upon FCTN decomposition~\cite{zhengFCTN} but can reveal the FCTN rank. It achieves this by inserting diagonal factors between any two TN cores in FCTN decomposition and leveraging the number of non-zero elements in the diagonal factors to determine the FCTN rank. In particular, SVDinsTN can transform into a TN decomposition by merging the diagonal factors into TN cores through the tensor contraction operation.
\end{remark}

\subsection{A Regularized Method for TN-SS} 

We present an SVDinsTN-based regularized method, which updates TN cores and diagonal factors alternately, and imposes a sparse operator to induce the sparsity of diagonal factors to reveal a compact TN structure.

We consider an $\ell_1$-norm-based operator for diagonal factors $\mathbf{S}$ and Tikhonov regularization~\cite{golub1999tikhonov} for TN cores $\mathcal{G}$. The $\ell_1$-norm-based operator is used to promote the sparsity of $\mathbf{S}$, and the Tikhonov regularization is used to constrict the feasible range of $\mathcal{G}$. Mathematically, the proposed model can be formulated as follows:
\begin{equation}
	\begin{aligned}
		\min_{\mathcal{G},\mathbf{S}}~&\frac{1}{2}\|\mathcal{X}-\text{STN}(\mathcal{G},\mathbf{S})\|_F^2+ \frac{\mu}{2}\sum\nolimits_{k\in\mathbb{K}_{N}}\|\mathcal{G}_k\|_F^2\\
		&+\sum\nolimits_{(t,l)\in\mathbb{TL}_{N}}\lambda_{t,l}\|\mathbf{S}_{t,l}\|_{1}, \\
	\end{aligned} \label{main_modelX2}
\end{equation}
where $\lambda_{t,l}>0$ and $\mu>0$ are regularization parameters.

We use the PAM-based algorithm~\cite{PAM_KL} to solve (\ref{main_modelX2}), whose solution is obtained by alternately updating
\begin{equation}
	\left\{
	\begin{aligned}
		&\mathcal{G}_k\!=\!\argmin_{\mathcal{G}_k}~\frac{1}{2}\|\mathcal{X} \!-\! \text{STN}(\mathcal{G},\mathbf{S})\|_F^2\!+\!\frac{\mu}{2}\|\mathcal{G}_k\|_F^2\\
		&~~~~~~~~~\!+\!\frac{\rho}{2}\|\mathcal{G}_k\!-\!\hat{\mathcal{G}}_k\|_F^2,\ \forall k\in\mathbb{K}_{N},\\
		&\mathbf{S}_{t,l}\!=\!\argmin_{\mathbf{S}_{t,l}}~\frac{1}{2}\|\mathcal{X} \!-\! \text{STN}(\mathcal{G},\mathbf{S})\|_F^2\!+\! \lambda_{t,l}\|\mathbf{S}_{t,l}\|_{1}\\
		&~~~~~~~~~\!+\!\frac{\rho}{2}\|\mathbf{S}_{t,l}\!-\!\hat{\mathbf{S}}_{t,l}\|_F^2,\ \forall (t,l)\in\mathbb{TL}_{N},\\
	\end{aligned}
	\right.
	\label{solve}
\end{equation}
where $\rho\!>\!0$ is a proximal parameter (we fix $\rho\!=\!0.001$), and $\hat{\mathcal{G}}_k$ and $\hat{\mathbf{S}}_{t,l}$ are the solutions of the $\mathcal{G}_k$-subproblem and $\mathbf{S}_{t,l}$-subproblem at the previous iteration, respectively.

\begin{algorithm}[!t]
	\small
	\caption{\small{$\mathbf{M}_k=\text{STN}\big(\{\mathcal{G}_q\}_{q=1}^N,\{\mathbf{S}_{t,l}\}_{1\leq t<l\leq N}^{t,l\in \mathbb{N}},/\mathcal{G}_k\big)$.}}
	\begin{algorithmic}[1]
			\renewcommand{\algorithmicrequire}{\textbf{Input:}}
			\Require
			$\mathcal{G}_q\in \mathbb{R}^{R_{1,q}\times R_{2,q}\times \cdots\times R_{q-1,q}\times I_q\times R_{q,q+1}\times\cdots\times R_{q,N}}$ for $\forall q\in\mathbb{K}_{N}$ and $q\neq k$; $\mathbf{S}_{t,l}\in\mathbb{R}^{R_{t,l}\times R_{t,l}}$ for $\forall(t,l)\in\mathbb{TL}_{N}$; and an index $k\in\mathbb{K}_{N}$.\vspace{0.05cm}
			\renewcommand{\algorithmicrequire}{\textbf{Initialization:}}
			\Require
			$\mathbf{a} = (k+1:N,1:k)$.
			\For {$i=1$ to $k-1$ and $i=k+1$ to $N-1$}
			\For {$j=i+1$ to $N$}
			\State Let $\mathcal{G}_i = \mathcal{G}_i\times_{i+1}^{1} \mathbf{S}_{i,j}$.
			\EndFor
			\If  {$i>k$}
			\State Let $\mathcal{G}_i = \mathcal{G}_i\times_{k}^{1} \mathbf{S}_{k,i}$.
			\State Let  $\mathcal{G}_i = \mathrm{permute}(\mathcal{G}_i,(1:k-1,N,k:N-1))$.
			\EndIf
			\State Let $\mathcal{G}_i = \mathrm{permute}(\mathcal{G}_i,\mathbf{a})$.
			\EndFor
			\State Let $\mathcal{M}_k=\mathcal{G}_{\mathbf{a}(1)}$, $m_1=1$, and $n_1=2$.
			\For {$i=1$ to $N-2$}
			\State Let $\mathcal{M}_k = \mathcal{M}_k\times_{n_1,n_2,\cdots,n_i}^{m_1,m_2,\cdots,m_i}\mathcal{G}_{\mathbf{a}(i+1)}$.
			\State Let $m_j=j$ for $j=1,2,\cdots,i+1$.
			\State Let $n_j=2+(j-1)(N-i)$ for $j=1,2,\cdots,i+1$.
			\EndFor
			\State Let $\mathcal{M}_k=\mathrm{permute}(\mathcal{M}_k,(2(N\!-\!k)+1:2(N-1),1:2(N-k)))$.
			\State Let $\mathbf{c}= \mathrm{zeros}(1,N-1)$ and $\mathbf{d}=\mathrm{zeros}(1,N-1)$.
			\For {$i=i$ to $N-1$}
			\State Let $\mathbf{c}(i)=2i$ and $\mathbf{d}(i)=2i-1$
			\EndFor
			\State Let $\mathbf{M}_k=\mathrm{GUnfold}(\mathcal{M}_k,\mathbf{c};\mathbf{d})$.\vspace{0.05cm}
			\renewcommand{\algorithmicrequire}{\textbf{Output:}}
			\Require Matrix $\mathbf{M}_k\in\mathbb{R}^{\prod_{i=1}^{k-1}R_{i,k}\prod_{i=k+1}^{N}R_{k,i}\times\prod_{i=1,i\neq k}^NI_i}$.
		\end{algorithmic}\label{GetXbyGSG}
\end{algorithm}

\emph{1) Update $\mathcal{G}_{k}$ for $\forall k\in\mathbb{K}_{N}$:} Solving the $\mathcal{G}_k$-subproblem requires fixing the other TN cores and diagonal factors. To address this, we use $\mathbf{M}_k$ to denote the matrix obtained by performing tensor contraction and unfolding operations on all diagonal factors and TN cores except $\mathcal{G}_k$. Algorithm~\ref{GetXbyGSG} presents a way to compute $\mathbf{M}_k$. We can obtain $\mathbf{X}_{(k)} = {\mathbf{G}_k}_{(k)}\mathbf{M}_{k}$. In this way, the $\mathcal{G}_{k}$-subproblem can be rewritten as follows:
\begin{equation}
	\begin{aligned}
		\min_{{\mathbf{G}_k}_{(k)}}~&\frac{1}{2}\|\mathbf{X}_{(k)}- {\mathbf{G}_k}_{(k)}\mathbf{M}_{k}\|_F^2+\frac{\mu}{2}\|{\mathbf{G}_k}_{(k)}\|_F^2\\
		&+\frac{\rho}{2}\|{\mathbf{G}_k}_{(k)}-\hat{\mathbf{G}}_k{}_{(k)}\|_F^2.
		\label{solveG}
	\end{aligned}
\end{equation}
The objective function of (\ref{solveG}) is differentiable, and thus its solution can be obtained by
\begin{equation}
	\begin{aligned}    {\mathbf{G}_k}_{(k)}\!=\!\big(\mathbf{X}_{(k)}\mathbf{M}_{k}^{\text{T}}\!+\!\rho\hat{\mathbf{G}}_k{}_{(k)}\big)
		\big(\mathbf{M}_{k}\mathbf{M}_{k}^{\text{T}}\!+\!(\mu\!+\!\rho)\mathbf{I}\big)^{-1}.
		\label{solveG2}
	\end{aligned}
\end{equation}

\emph{2) Update $\mathbf{S}_{t,l}$ for $\forall (t,l)\in\mathbb{TL}_{N}$:} Solving the $\mathbf{S}_{t,l}$-subproblem requires fixing the other diagonal factors and TN cores. In a similar fashion, we use $\mathbf{H}_{t,l}$ to denote the matrix obtained by performing tensor contraction and unfolding operations on all TN cores and diagonal factors except $\mathbf{S}_{t,l}$. Algorithm~\ref{GetXbyGSS} presents a way to compute $\mathbf{H}_{t,l}$. We can obtain $\mathbf{x} = \mathbf{H}_{t,l}\mathbf{s}_{t,l}$, where $\mathbf{x}=\mathrm{vec}(\mathcal{X})$ and $\mathbf{s}_{t,l}=\mathrm{diag}(\mathbf{S}_{t,l})$. 
Then, the $\mathbf{S}_{t,l}$-subproblem can be rewritten as follows:
\begin{equation}
	\begin{aligned}
		\min_{\mathbf{s}_{t,l}}&~\frac{1}{2}\|\mathbf{x}\!-\!\!\mathbf{H}_{t,l}\mathbf{s}_{t,l}\|_F^2\!+\!\lambda_{t,l}\|\mathbf{s}_{t,l}\|_{1}
		\!\!+\!\frac{\rho}{2}\|\mathbf{s}_{t,l}\!\!-\!\hat{\mathbf{s}}_{t,l}\|_F^2.
		\label{solveS}
	\end{aligned}
\end{equation}

\begin{algorithm}[!t]
	\small
	\caption{\small{$\mathbf{H}_{t,l}=\text{STN}\big(\!\{\mathcal{G}_k\}_{k=1}^N,\{\mathbf{S}_{p,q}\}_{1\leq p<q\leq N}^{p,q\in \mathbb{N}},/\mathbf{S}_{t,l}\big)$.}}
	\begin{algorithmic}[1]
		\renewcommand{\algorithmicrequire}{\textbf{Input:}}
		\Require
		$\mathcal{G}_k\in \mathbb{R}^{R_{1,k}\times R_{2,k}\times \cdots\times R_{k-1,k}\times I_k\times R_{k,k+1}\times\cdots\times R_{k,N}}$ for $\forall k\in\mathbb{K}_{N}$; $\mathbf{S}_{p,q}\in\mathbb{R}^{R_{p,q}\times R_{p,q}}$ for $\forall(p,q)\in\mathbb{TL}_{N}$, and $(p,q)\neq (t,l)$; and an index $(t,l)\in\mathbb{TL}_{N}$.\vspace{0.05cm}
		\For {$i=1$ to $t-1$ and $i=t+1$ to $N-1$}
		\For {$j=i+1$ to $N$}
		\State Let $\mathcal{G}_i = \mathcal{G}_i\times_{i+1}^{1} \mathbf{S}_{i,j}$.
		\EndFor
		\EndFor
		\For {$j=t+1$ to $l-1$}
		\State Let $\mathcal{G}_t = \mathcal{G}_t\times_{t+1}^{1} \mathbf{S}_{t,j}$.
		\EndFor
		\For {$j=l+1$ to $N$}
		\State Let $\mathcal{G}_t = \mathcal{G}_t\times_{t+2}^{1} \mathbf{S}_{t,j}$.
		\EndFor
		\State Let $\mathcal{G}_t=\mathrm{permute}(\mathcal{G}_t,(1:t,t+2:l,t+1,l+1:N))$.
		\State Let $\mathbf{G}_t= \mathrm{Unfold}(\mathcal{G}_t,l)$ and $\mathbf{G}_l= \mathrm{Unfold}(\mathcal{G}_l,t)$.

		\State Let $\mathbf{H}_{t,l}=\mathrm{zeros}(\prod_{k=1}^NI_k, R_{t,l})$.
		\For {$i=1$ to $R_{t,l}$}
		\State Let $\mathcal{G}_t= \mathrm{Fold}(\mathbf{G}_t(i,:),l)$ and $\mathcal{G}_l= \mathrm{Fold}(\mathbf{G}_l(i,:),t)$.
		\State Let $\mathbf{H}_{t,l}(:,i)=\mathrm{vec}(\text{FCTN}(\{\mathcal{G}_k\}_{k=1}^N))$.
		\EndFor\vspace{0.05cm}
		\renewcommand{\algorithmicrequire}{\textbf{Output:}}
		\Require Matrix $\mathbf{H}_{t,l}\in\mathbb{R}^{\prod_{k=1}^NI_k\times R_{t,l}}$.
	\end{algorithmic}\label{GetXbyGSS}
\end{algorithm}

We use an alternating direction method of multipliers (ADMM)~\cite{GABAY197617} to solve the $\mathbf{S}_{t,l}$-subproblem, which can be rewritten as follows:
\begin{equation}
	\begin{aligned}
		\min_{\mathbf{s}_{t,l},\mathbf{q}_{t,l}}&~\frac{1}{2}\|\mathbf{x}\!-\!\!\mathbf{H}_{t,l}\mathbf{q}_{t,l}\|_F^2\!+\!\lambda_{t,l}\|\mathbf{s}_{t,l}\|_{1}
		\!\!+\!\frac{\rho}{2}\|\mathbf{s}_{t,l}\!\!-\!\hat{\mathbf{s}}_{t,l}\|_F^2\\
		\text{s.t.}&~\mathbf{s}_{t,l}\!-\!\mathbf{q}_{t,l}\!=\!0,
		\label{solveS2}
	\end{aligned}
\end{equation}
where $\mathbf{q}_{t,l}$ is an auxiliary variable.
The augmented Lagrangian function of (\ref{solveS2}) can be expressed as the following concise form: 
\begin{equation}
	\begin{aligned}
		L_{\beta_{t,l}}&(\mathbf{s}_{t,l},\mathbf{q}_{t,l},\mathbf{p}_{t,l})\!=\!\frac{1}{2}\|\mathbf{x}\!-\!\mathbf{H}_{t,l}\mathbf{q}_{t,l}\|_F^2\!+\!\lambda_{t,l}\|\mathbf{s}_{t,l}\|_{1}\\
		&\!+\!\frac{\rho}{2}\|\mathbf{s}_{t,l}\!-\!\hat{\mathbf{s}}_{t,l}\|_F^2 \!+\!\frac{\beta_{t,l}}{2}\Big\|\mathbf{s}_{t,l}\!-\!\mathbf{q}_{t,l}\!+\!\frac{\mathbf{p}_{t,l}}{\beta_{t,l}}\Big\|_F^2,
		\label{solveSL}
	\end{aligned}
\end{equation}
where $\mathbf{p}_{t,l}$ is the Lagrangian multiplier and $\beta_{t,l}>0$ is the penalty parameter. Within the ADMM framework, $\mathbf{q}_{t,l}$, $\mathbf{s}_{t,l}$, and $\mathbf{p}_{t,l}$ can be solved by alternately updating
\begin{equation}
	\left\{
	\begin{aligned}
		&\mathbf{q}_{t,l}=\argmin_{\mathbf{q}_{t,l}}~L_{\beta_{t,l}}(\mathbf{s}_{t,l},\mathbf{q}_{t,l},\mathbf{p}_{t,l}),\\
		&\mathbf{s}_{t,l}=\argmin_{\mathbf{s}_{t,l}}~L_{\beta_{t,l}}(\mathbf{s}_{t,l},\mathbf{q}_{t,l},\mathbf{p}_{t,l}),\\
		&\mathbf{p}_{t,l} = \mathbf{p}_{t,l}+\beta_{t,l}(\mathbf{s}_{t,l}-\mathbf{q}_{t,l}).
	\end{aligned}
	\right.
	\label{solveSADMM0}
\end{equation}
That is, 
\begin{equation}
	\left\{
	\begin{aligned}
		&\mathbf{q}_{t,l}\!=\!\big[\mathbf{H}_{t,l}^{\text{T}}\mathbf{H}_{t,l}\!+\!\beta_{t,l}\mathbf{I}\big]^{-1}\big[\mathbf{H}_{t,l}^{\text{T}}\mathbf{x} \!+\!\beta_{t,l}\mathbf{s}_{t,l}\!+\!\mathbf{p}_{t,l}\big],\\
		&\mathbf{s}_{t,l}\!=\!\mathrm{ shrink}\bigg(\frac{\rho\hat{\mathbf{s}}_{t,l}\!+\!\beta_{t,l}\mathbf{q}_{t,l}\!-\!\mathbf{p}_{t,l}}{\rho\!+\!\beta_{t,l}},\frac{\lambda_{t,l}}{\rho\!+\!\beta_{t,l}}\bigg),\\
		&\mathbf{p}_{t,l}\! =\! \mathbf{p}_{t,l}\!+\!\beta_{t,l}(\mathbf{s}_{t,l}\!-\!\mathbf{q}_{t,l}),
	\end{aligned}
	\right.
	\label{solveSADMM}
\end{equation}
where $\mathrm{shrink}(\mathbf{a},\mathbf{b})=\mathrm{max}(\mathbf{a}-\mathbf{b},\mathbf{0})+\mathrm{min}(\mathbf{a}+\mathbf{b},\mathbf{0})$.

\begin{algorithm}[!t]
	\small
	\caption{PAM-based algorithm to optimize model~(\ref{main_modelX2}).}
	\begin{algorithmic}[1]
		\renewcommand{\algorithmicrequire}{\textbf{Input:}}
		\Require 
		A tensor $\mathcal{X}\in\mathbb{R}^{I_1\times I_2\times\cdots\times I_N}$ and a parameter $\gamma$.\vspace{0.05cm}
		\renewcommand{\algorithmicrequire}{\textbf{Initialization:}}
		\Require Initialize $\mathbf{S}_{t,l}$ and $R_{t,l}$ by the initialization scheme in Section~\ref{Initialization} and let $\beta_{t,l}=1$ for $\forall (t,l)\in\mathbb{TL}_{N}$; let
		$\mathcal{G}_{k}=1/\sqrt{I_k}\,\mathrm{ones}(R_{1,k},R_{2,k},\cdots,R_{k-1,k},I_k,R_{k,k+1},\cdots, R_{k,N})$ for $\forall k\in\mathbb{K}_{N}$ and $\mu=1$.
		\While {not converged}
		\State Let $\hat{\mathcal{X}}=\mathcal{X}$ and $\lambda_{t,l}=\gamma\,\mathrm{max}(\mathbf{S}_{t,l})(\rho+\beta_{t,l})$.
		\State Update $\mathbf{G}_k{}_{(k)}$ by (\ref{solveG2}) and let $\mathcal{G}_k=\mathrm{Fold}(\mathbf{G}_k{}_{(k)}, k)$.
		\For {$i=1$ to $5$}
		\State Update $\mathbf{q}_{t,l}$, $\mathbf{s}_{t,l}$, and $\mathbf{p}_{t,l}$ by (\ref{solveSADMM}).
		\EndFor
		\State Delete zero elements in $\mathbf{s}_{t,l}$, let $\mathbf{S}_{t,l}=\mathrm{diag}(\mathbf{s}_{t,l})$, and define the size of $\mathbf{s}_{t,l}$ as $R_{t,l}$.
		\State Delete the corresponding dimensions of $\mathcal{G}_{k}$ and let $\mathcal{X}= \text{STN}(\mathcal{G},\mathbf{S})$. 
		\State Check the convergence condition: 
		$\frac{\|\mathcal{X}-\hat{\mathcal{X}}\|_{F}}{\|\hat{\mathcal{X}}\|_{F}}<10^{-5}$.
		\EndWhile
		\vspace{0.05cm}
		\renewcommand{\algorithmicrequire}{\textbf{Output:}}
		\Require $\mathcal{G}_{k}$ for $\forall k\in\mathbb{K}_{N}$, and $\mathbf{S}_{t,l}$ and $R_{t,l}$ for $\forall (t,l)\in\mathbb{TL}_{N}$.
	\end{algorithmic}\label{AlgforSTNm}
\end{algorithm}

We describe the pseudocode to optimize model~(\ref{main_modelX2}) in Algorithm \ref{AlgforSTNm}. Below, we present a brief analysis of the computational complexity and provide a theoretical convergence guarantee for the developed algorithm.

\textbf{Computational complexity.}  For simplicity, we let the size of the $N$th-order tensor $\mathcal{X}$ be $I\times I\times \cdots\times I$ and the initial rank be $(R,R,\cdots,R)$ satisfied $R\leq I$. The computational cost involves updating $\mathcal{G}$ and $\mathbf{S}$, resulting in costs of $\mathcal{O}\big(N\sum_{k=2}^{N}I^kR^{k(N-k)+k-1}\!+\!NI^{N-1}R^{2(N-1)}\!+\!N^3IR^N\big)$ and $\mathcal{O}\big(N^2\sum_{k=2}^{N}I^kR^{k(N-k)+k-1}\!+\!N^4IR^N\!+\!N^2I^{N}R^2\big)$, respectively. Hence, the computational cost at each iteration is $\mathcal{O}\big(N^2\sum_{k=2}^{N}I^kR^{k(N-k)+k-1}\!+\!N^4IR^N\!+\!N^2I^{N}R^2\big)$.

\begin{theorem}[Convergence guarantee]\label{Conver}
	The sequence generated by Algorithm \ref{AlgforSTNm}, denoted by $\{\mathcal{G}^{(s)},\mathbf{S}^{(s)}\}_{s\in{\mathbb{N}}}$,  converges to a critical point of the optimization problem (\ref{main_modelX2}).
\end{theorem}

\subsection{Initialization Scheme}
\label{Initialization}

SVDinsTN encounters high computational cost in the first several iterations if the TN rank $R_{t,l}$ for $\forall (t,l)\in\mathbb{TL}_{N}$ are initialized with large values. This is due to the adoption of a ``fully-connected'' topology as a starting point. To solve this challenge, we design a novel initialization scheme aimed at effectively reducing the initial values of the TN rank.

We first give an upper bound for the TN rank, by which we then design an initialization scheme for the TN rank $R_{t,l}$ and diagonal factors $\mathbf{S}_{t,l}\in \mathbb{R}^{R_{t,l}\times R_{t,l}}$ for $\forall (t,l)\in\mathbb{TL}_{N}$.
\begin{theorem}\label{FCTNDt} Let $\mathcal{X}\in\mathbb{R}^{I_1\times I_2\times\cdots\times I_N}$ be an $N$th-order tensor, then there exists an SVDinsTN (\ref{TNtensor3}) with the TN rank ${R}_{t,l}\leq \mathrm{min}(\mathrm{rank}(\mathbf{X}_{(t)}), \mathrm{rank}(\mathbf{X}_{(l)}))$ for $\forall (t,l)\in\mathbb{TL}_{N}$.
\end{theorem}

Theorem~\ref{FCTNDt} indicates that $\mathrm{min}(\mathrm{rank}(\mathbf{X}_{(t)}),\mathrm{rank}(\mathbf{X}_{(l)}))$ can be the initial value of the TN rank ${R}_{t,l}$. For real-world data, this value is usually embodied by the rank of mode-$(t,l)$ slices\footnote{Mode-$(t,l)$ slices are obtained by fixing all but the mode-$t$ and the mode-$l$ indexes of a tensor~\cite{IS_Ntubal}.} of $\mathcal{X}$. Therefore, we initialize $R_{t,l}$ and $\mathbf{S}_{t,l}$ by virtue of truncated SVD of mode-$(t,l)$ slices of $\mathcal{X}$, which consists of the following two steps.

\textit{Step 1:} We first calculate the mean of all mode-$(t,l)$ slices of $\mathcal{X}$ and denote it by $\mathbf{X}_{t,l}$. Then we perform SVD on $\mathbf{X}_{t,l}$ to obtain $\mathbf{s}_{t,l}\in \mathbb{R}^{\mathrm{min}(I_t, I_l)}$, whose elements are singular values of $\mathbf{X}_{t,l}$.

\textit{Step 2:} We first let $\mathbf{s}_{t,l}=\mathrm{shrink}\big(\mathbf{s}_{t,l},\frac{\gamma\,\mathrm{ max}(\mathbf{s}_{t,l})}{|\mathbf{s}_{t,l}|+10^{-16}}\big)$ and delete zero elements in $\mathbf{s}_{t,l}$. Then we let $\mathbf{S}_{t,l}=\mathrm{diag}(\mathbf{s}_{t,l})$ and define the size of $\mathbf{s}_{t,l}$ as $R_{t,l}$.

In practical applications, the $\mathrm{shrink}$ operation in Step 2 effectively reduces the initial value of $R_{t,l}$ by projecting very small singular values in $\mathbf{s}_{t,l}$ to zero. As a result, \textit{the challenge of high computational costs in the first several iterations of SVDinsTN can be effectively addressed} (see Figure~\ref{illu_int} for a numerical illustration).

\begin{table*}[!t]
	\setlength{\tabcolsep}{12pt}
	\small
	\renewcommand\arraystretch{1}
	\caption{Performance of SVDinsTN on TN structure revealing under $100$ independent tests.}
	\label{symfivetab}
	\centering
	\begin{tabular}{cccccc}
		\toprule
		
		{True structure (4th-order)} &  			\raisebox{-.46\height}{\includegraphics[width=0.1\textwidth]{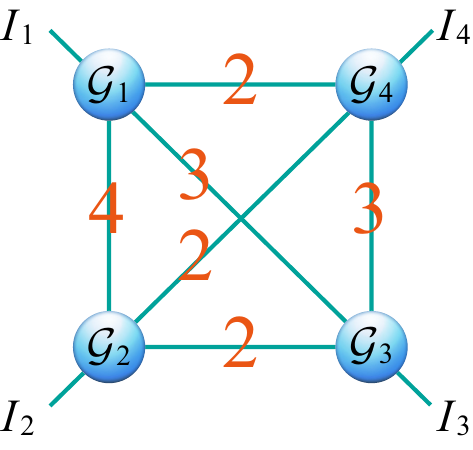}}
		&  \raisebox{-.46\height}{\includegraphics[width=0.1\textwidth]{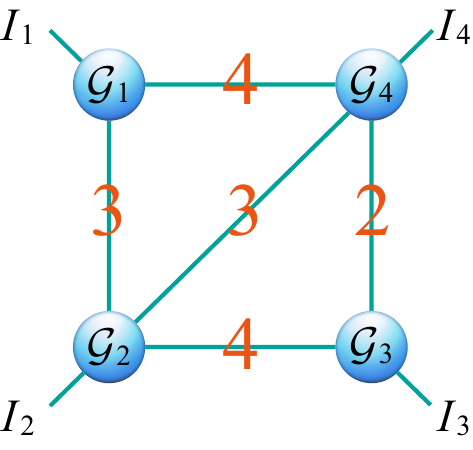}}
		&  \raisebox{-.46\height}{\includegraphics[width=0.1\textwidth]{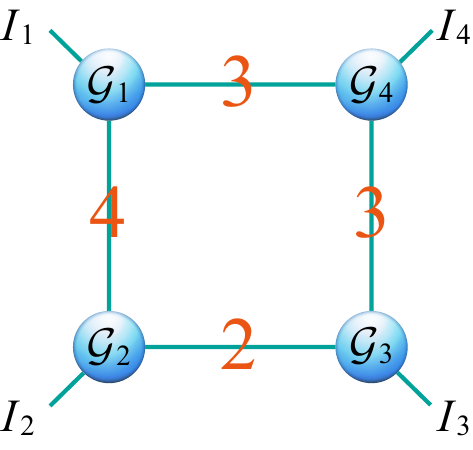}} 
		&  \raisebox{-.46\height}{\includegraphics[width=0.1\textwidth]{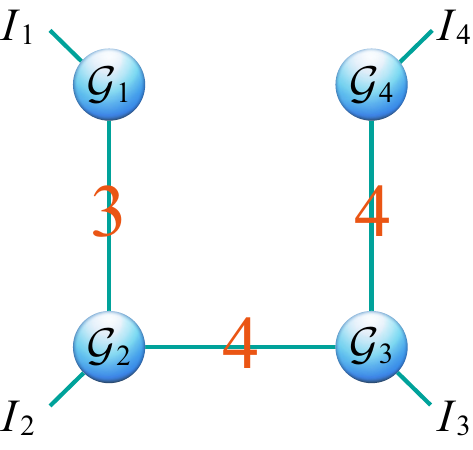}}
		&  \raisebox{-.46\height}{\includegraphics[width=0.1\textwidth]{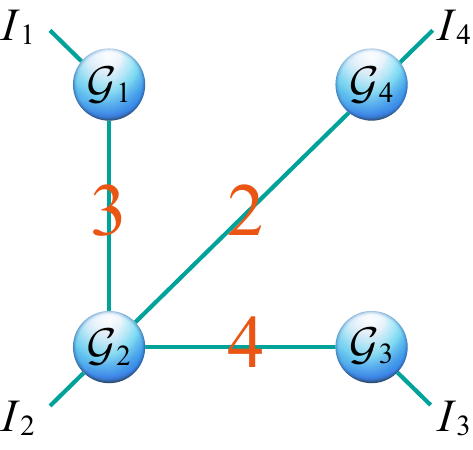}}\\
		
		\midrule
		
		Success rate  & 100\%  &  100\% &   96\% & 95\%   &99\%\\[-0.05cm]
		\midrule
		
		{True structure (5th-order)} &  \raisebox{-.46\height}{\includegraphics[width=0.1\textwidth]{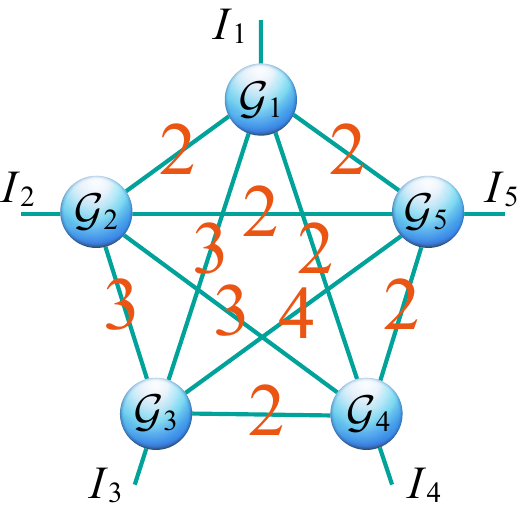}}
		&  \raisebox{-.46\height}{\includegraphics[width=0.1\textwidth]{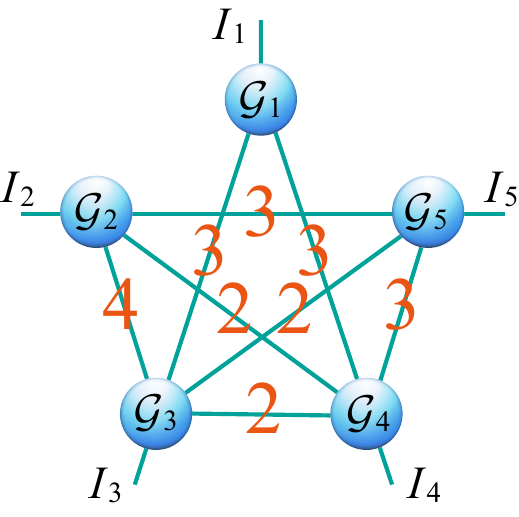}}
		&  \raisebox{-.46\height}{\includegraphics[width=0.1\textwidth]{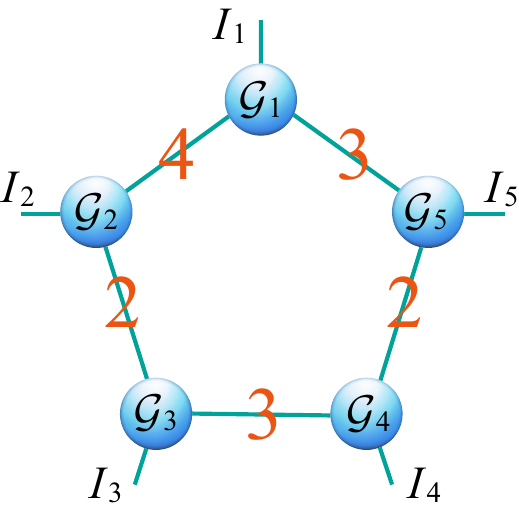}}  
		&  \raisebox{-.46\height}{\includegraphics[width=0.1\textwidth]{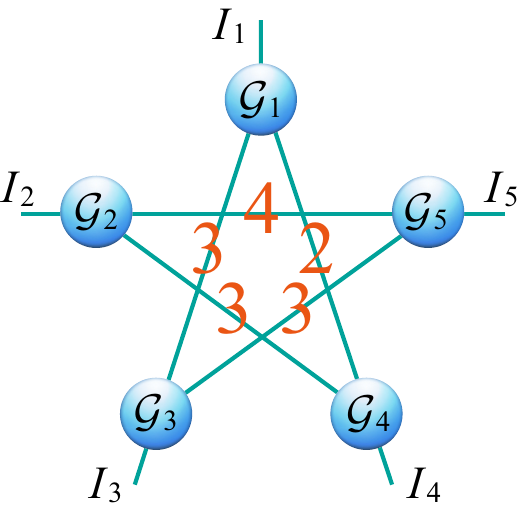}}
		&  \raisebox{-.46\height}{\includegraphics[width=0.1\textwidth]{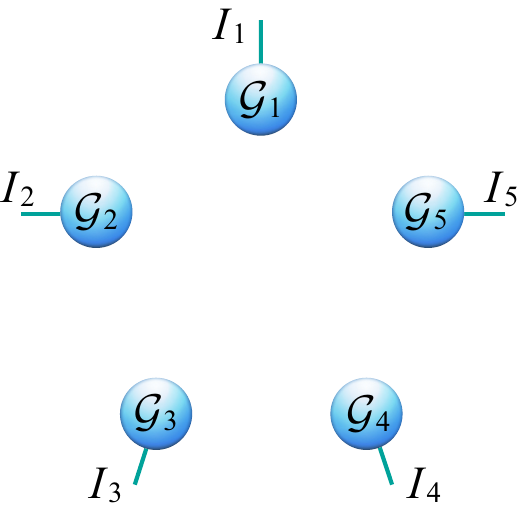}}\\
		
		\cmidrule{1-6}
		
		{Success rate}  & 100\%  &  98\% &   96\% & 97\%   &100\%\\
		\bottomrule
	\end{tabular}\vspace{-0.0cm}
\end{table*}

\section{Numerical Experiments} \label{NumericalExper}

In this section, we present numerical experiments on both synthetic and real-world data to evaluate the performance of the proposed SVDinsTN. The primary objective is to validate the following three Claims:

\begin{enumerate}
	\item[A:] SVDinsTN can reveal a customized TN structure that aligns with the unique structure of a given tensor.
	\item[B:] SVDinsTN can greatly reduce time costs while achieving a comparable representation ability to state-of-the-art TN-SS methods. Moreover, SVDinsTN can also surpass existing tensor decomposition methods with pre-defined topologies regarding representation ability.
	\item[C:] SVDinsTN can outperform existing tensor decomposition methods in the tensor completion task, highlighting its effectiveness as a valuable tool in applications.
\end{enumerate}

\subsection{Experiments for Validating Claim A}\label{secClaimA}

We conduct experiments to validate Claim A. Since real-world data lacks a true TN structure, we consider only synthetic data in this experiment. 

\textbf{Data generation.} We first randomly generate $\mathcal{G}_k$ for $\forall k\in\mathbb{K}_{N}$ and $\mathbf{S}_{t,l}$ for $\forall (t,l)\in\mathbb{TL}_{N}$, whose elements are taken from a uniform distribution between 0 and 1. Then we obtain the synthetic tensor by $\mathcal{X}= \text{STN}(\mathcal{G},\mathbf{S})$.

\textbf{Experiment setting.} We test both fourth-order tensors of size $16\times 18\times 20\times 22$ and fifth-order tensors of size $14\times 16\times 18\times 20\times 22$, and consider different kinds of TN structures. For each structure, we conduct $100$ independent tests and regenerate the synthetic data to ensure reliable and unbiased results. The ability of the proposed SVDinsTN to reveal TN structure is measured by the \textit{success rate} of the output structures, defined as ${S_{\text{T}}}/{T}\times100\%$, where $T=100$ is the total number of tests and $S_{\text{T}}$ is the number of tests that accurately output the true TN structure. In all tests, the parameter $\gamma$ is set to $0.0015$.

Table~\ref{symfivetab} presents the \textit{success rate} of the output TN structures obtained by the proposed SVDinsTN in $100$ independent tests on fourth-order and fifth-order tensors. It can be observed that the proposed SVDinsTN consistently yields high \textit{success rates} of over $95\%$ in all test cases. Notably, in approximately half of the test cases, the \textit{success rates} reach a perfect score of $100\%$. Moreover, it is also worth mentioning that in the test on fifth-order tensors, we consider two isomorphic topologies: the ``ring'' topology and the ``five-star'' topology. These two topologies are both the ``ring'' topology (TR decomposition), but with different permutations: $\mathcal{G}_1\!\rightarrow\!\mathcal{G}_2\!\rightarrow\!\mathcal{G}_3\!\rightarrow\!\mathcal{G}_4\!\rightarrow\!\mathcal{G}_5\!\rightarrow\!\mathcal{G}_1$  and $\mathcal{G}_1\!\rightarrow\!\mathcal{G}_3\!\rightarrow\!\mathcal{G}_5\!\rightarrow\!\mathcal{G}_2\!\rightarrow\!\mathcal{G}_4\!\rightarrow\!\mathcal{G}_1$, respectively. It can be seen that despite the isomorphism, the proposed SVDinsTN can identify the correct permutation for each topology.

\subsection{Experiments for Validating Claim B} \label{seccr}

We conduct experiments to validate Claim B. We consider both real-world data and synthetic data, and use different methods to represent it in this experiment.

\begin{table}[!t]
	\small
	%\tiny
	\setlength{\tabcolsep}{1.6pt}
	\renewcommand\arraystretch{1}
	\caption{Comparison of CR ($\downarrow$) and run time ($\times$1000s, $\downarrow$) of different methods on light field data.}
	\label{FLpertab}
	\centering
	\begin{tabular}{ccccccc}
		\toprule
		\multirow{2}{*}{Method} & \multicolumn{2}{c}{RE~bound:~0.01}  & \multicolumn{2}{c}{RE~bound:~0.05}   & \multicolumn{2}{c}{RE~bound:~0.1}              \\
		\cmidrule(rl){2-3}\cmidrule(rl){4-5} \cmidrule(rl){6-7}
		&  CR  & Time &  CR & Time & CR  & Time \\
		
		\midrule\midrule
		\multicolumn{7}{c}{\textit{Bunny}}\\
		\midrule
		
		TRALS~\cite{zhao2016tensor} & 60.5\%  & 13.54 & 17.4\%  & 0.471 & 5.31\%  & 0.118\\
		FCTNALS~\cite{zhengFCTN}  &  65.1\% & 13.08	& 20.9\%   &  0.473	& 3.93\%   &  0.041	\\
		TNGreedy~\cite{hashemizadeh2022adaptive} & 26.1\% & 11.02	& 6.32\%  & 1.021	&2.34\%  &0.362\\
	    TNGA~\cite{LiTopologySearch} & 27.9\% & 1014&{5.01\%} &180.3 	& {2.25\%}   & 12.52\\
		%TNGA+~\cite{LiTopologySearch} & \underline{15.0\%} &2098 &{4.48\%} &248.7	& \underline{2.13\%}   &16.29\\
		TNLS~\cite{LiPermutationSearch} &  24.3\% &1402&	\underline{4.26\%} &63.70&	\underline{2.16\%}&24.53\\
		TNALE~\cite{li2023alternating} &  26.3\% &144.5&	4.52\% &18.36&	2.26\%&3.064\\
		SVDinsTN  &  \underline{22.4\%} &\textbf{0.745}	& 6.92\%	&	\textbf{0.029}	& 2.66\%	& \textbf{0.005}\\	
		
		\midrule\midrule
		\multicolumn{7}{c}{\textit{Knights}}\\
		\midrule

		TRALS~\cite{zhao2016tensor} & 74.7\%  & 10.31& 26.9\%  & 3.835 & 9.15\%  & 0.423	\\
		FCTNALS~\cite{zhengFCTN}  & 73.5\%  & 12.35& 20.9\%  & 0.619& 3.93\%   & \textbf{0.014}\\
		TNGreedy~\cite{hashemizadeh2022adaptive}  & 32.1\% & 12.53	& 7.55\%  & 1.366	&3.50\%  &0.481\\
		TNGA~\cite{LiTopologySearch} & 38.7\% & 912.9 	& {5.01\%} &140.2 	& {2.44\%}    & 12.52\\
		%TNGA+~\cite{LiTopologySearch} & \underline{19.6\%} & 2889&{4.55\%} &308.4	& {2.24\%}   &13.16\\
		TNLS~\cite{LiPermutationSearch} &  \underline{27.3\%} &1286&	4.73\% &75.51&	2.15\%&5.320\\
		TNALE~\cite{li2023alternating} &  {27.6\%} &266.4&	\underline{4.52\%} &25.05&	\underline{2.10\%}&3.386\\
		SVDinsTN  & 32.0\% &\textbf{1.548}& 5.64\% &\textbf{0.104}	& 2.76\% & 0.019	\\	
		
		\midrule\midrule
		\multicolumn{7}{c}{\textit{Truck}}\\
		\midrule

		TRALS~\cite{zhao2016tensor}  & 62.8\% & 17.62 & 22.6\%   & 1.738	& 6.00\%   & 0.090	\\
		FCTNALS~\cite{zhengFCTN}  & 69.3\% & 7.735  & 20.9\%  & 2.953  & 3.93\%   & 0.159\\
		TNGreedy~\cite{hashemizadeh2022adaptive}  & 26.9\% & 6.676	& 7.26\%  & 1.259	&3.35\%  &0.488\\
		TNGA~\cite{LiTopologySearch}  & 27.9\%   & 1029  & {5.01\%} & 170.3  & 2.85\%  & 14.83 \\
	    %TNGA+~\cite{LiTopologySearch} & \underline{18.0\%} & 2433&\underline{3.87\%} &	189.1& \underline{2.29\%}   &5.638\\
	   	TNLS~\cite{LiPermutationSearch} &  26.4\% &992.6&	\underline{4.99\%} &119.8&	\underline{2.57\%}&19.35\\
		TNALE~\cite{li2023alternating} &  24.7\% &239.3&5.77\% &19.54&	2.90\%&5.160\\
		SVDinsTN  & \underline{23.5\%}  & \textbf{1.051} &6.42\% & \textbf{0.152}  &2.83\% & \textbf{0.023}\\	
		\bottomrule
	\end{tabular}
	\vspace{-0.3cm}
\end{table}

\textbf{Experiment setting.} We test three light field data\footnote{\url{http://lightfield.stanford.edu/lfs.html}}, named \textit{Bunny}, \textit{Knights}, and \textit{Truck}, which are fifth-order tensors of size $40\times 60\times 3\times 9\times 9$ (spatial height $\!\times\!$ spatial width $\!\times\!$ color channel $\!\times\!$ vertical grid $\!\times\!$ horizontal grid). We employ six representative methods as the compared baselines, including two methods with pre-defined topology: TRALS~\cite{zhao2016tensor} and FCTNALS~\cite{zhengFCTN}, and four TN-SS methods: TNGreedy~\cite{hashemizadeh2022adaptive}, TNGA~\cite{LiTopologySearch}, TNLS~\cite{LiPermutationSearch}, and TNALE~\cite{li2023alternating}. We represent the test light field data by different methods and calculate the corresponding \textit{compression ratio (CR)} to achieve a certain \textit{reconstruction error (RE) bound}. The \textit{CR} is defined as $F_{\mathcal{G}}/F_{\mathcal{X}}\times100\%$, where $F_{\mathcal{G}}$ is the number of elements of TN cores used to represent a tensor and $F_{\mathcal{X}}$ is the number of total elements of the original tensor. The \textit{RE} is defined as $\|\mathcal{X}-\tilde{\mathcal{X}}\|_F/\|\mathcal{X}\|_F$, where $\mathcal{X}$ is the original data and $\tilde{\mathcal{X}}$ is the reconstructed data. In all tests, we select the parameter $\gamma$ from the interval $[10^{-7},10^{-3}]$.

\textbf{Result analysis.} Table~\ref{FLpertab} reports \textit{CR} and \textit{run time} of different methods on fifth-order light field data. The results show that the proposed SVDinsTN achieves significantly lower \textit{CRs} than TRALS and FCTNALS, which are methods with pre-defined topology. This indicates that SVDinsTN can obtain a more compact structure than the pre-defined one. Furthermore, while SVDinsTN requires the determination of diagonal factors alongside TN cores, its iterative process generates progressively simpler structures, enhancing the computational efficiency. Consequently, SVDinsTN demonstrates \textit{faster} performance compared to TRALS and FCTNALS. Compared to the TN-SS methods, the proposed SVDinsTN achieves \textit{a substantial speed improvement} while maintaining a comparable level of \textit{CR}. Remarkably, SVDinsTN achieves an acceleration of approximately $100\!\sim{}\!1000$ times over TNGA, TNLS, and TNALE. This is because TNGreedy, TNGA, TNLS, and TNALE adopt the ``sampling-evaluation'' framework, necessitating a significant number of repeated structure evaluations. In contrast, SVDinsTN introduces a regularized modeling framework, requiring only a single evaluation.

\begin{table}[!t]
	\small
	%\tiny
	\setlength{\tabcolsep}{2.8pt}
	\renewcommand\arraystretch{1}
	\caption{Comparison of CR ($\downarrow$) and run time ($\times$1000s, $\downarrow$) of SVDinsTN with different initializations on light field data \textit{Truck}.}
	\label{chushitable}
	\centering
	\begin{tabular}{ccccccc}
		\toprule
		\multirow{2}{*}{Initialization} & \multicolumn{2}{c}{RE~bound:~0.01}  & \multicolumn{2}{c}{RE~bound:~0.05}   & \multicolumn{2}{c}{RE~bound:~0.1}              \\
		\cmidrule(rl){2-3}\cmidrule(rl){4-5} \cmidrule(rl){6-7}
		&  CR  & Time &  CR & Time & CR  & Time \\
		
		\midrule

		Random &  30.7\% &1.203&8.17\% &0.474&	4.14\%&0.208\\
		Ours  & 23.5\%  & 1.051 &6.42\% & 0.152  &2.83\% & 0.023\\
		\bottomrule
	\end{tabular}
\end{table}

\begin{figure}
	\centering
	\includegraphics[width=0.475\textwidth]{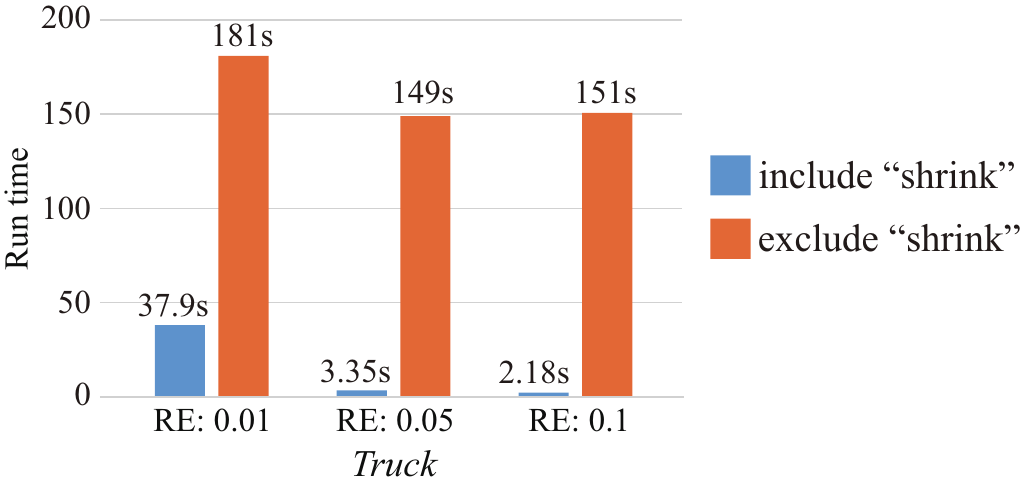}
	\caption{Comparison of the runtime in the first five iterations of SVDinsTN on light field data \textit{Truck} when including and excluding the $\mathrm{shrink}$ operation in our initialization scheme.}	\vspace{-0.3cm}
	\label{illu_int}
\end{figure}

\textbf{Impact of the initialization scheme.} We analyze the impact of the initialization scheme. In Table~\ref{chushitable}, we report \textit{CR} and \textit{run time} of SVDinsTN with different initializations on light field data \textit{Truck}. As observed, our initialization scheme achieves lower \textit{CRs} compared to random initialization, while maintaining higher efficiency. This corroborates that our initialization scheme can provide a favorable starting point and enhance computational efficiency. In particular, even with random initialization, our method achieves significant acceleration compared to other TN-SS methods. We further analyze the impact of the $\mathrm{shrink}$ operation in our initialization scheme. In Figure~\ref{illu_int}, we present the \textit{run time} comparison of the first five iterations of our method when including and excluding the $\mathrm{shrink}$ operation in our initialization scheme. As observed, the $\mathrm{shrink}$ operation in our initialization scheme enables our method to greatly reduce the computational costs in the first several iterations.% These findings validate the analysis presented in Section~\ref{Initialization}.

\textbf{Higher-order cases.} We analyze whether the proposed SVDinsTN still performs well on higher-order tensors. We randomly generate 6th-, 8th-, and 10th-order tensors by using the same procedure in Section~\ref{secClaimA}. The size of each tensor mode is randomly selected from $\{5, 6, 7, 8\}$, the edge number of each TN is randomly selected from $\{6, 8, 10\}$, and the rank of each edge is randomly selected from $\{2, 3\}$. For each tensor order, we randomly generate 5 tensors. We compare SVDinsTN and baseline methods in terms of \textit{CR} and \textit{run time} when
reaching the RE bound of 0.01, and show the results in Table~\ref{ordertab}. As observed, SVDinsTN is applicable to higher orders beyond 5, and even up to 10. The behind rational is the truncated SVD used in initialization restricts
the initial values of the rank for each edge to a relatively small range,
thus improving computational and storage efficiency (see Figure~\ref{illu_int}). As the iterations progress,
the sparsity regularization in the model leads to progressively simpler learned structures, further boosting efficiency.

\begin{table}[!t]
	\small
	\setlength{\tabcolsep}{2.4pt}
	\renewcommand\arraystretch{1}
	\caption{Comparison of the CR ($\downarrow$) and run time ($\times$1000s, $\downarrow$) of different methods when reaching the RE bound of 0.01. The result is the average value of 5 independent experiments and ``--'' indicates ``out of memory''.
	}
	\label{ordertab}
	\centering
	\begin{tabular}{ccccccc}
		\toprule
		\multirow{2}{*}{Method~} & \multicolumn{2}{c}{6th-order}  & \multicolumn{2}{c}{8th-order}   & \multicolumn{2}{c}{10th-order}              \\
		\cmidrule(rl){2-3}\cmidrule(rl){4-5} \cmidrule(rl){6-7}
		&  CR  & Time &  CR & Time & CR  & Time \\
		
		\midrule
		
		TRALS~\cite{zhao2016tensor} & 1.35\%  & 0.006 & 0.064\%  & 0.034 & --  & --\\
		FCTNALS~\cite{zhengFCTN}  & 2.13\% & \textbf{0.002}	& --   &  --	& --   &  --	\\
		TNGreedy~\cite{hashemizadeh2022adaptive} & \underline{0.88\%} & 0.167	& \underline{0.016\%}  & 2.625	&0.0008\%  &45.39\\
		TNGA~\cite{LiTopologySearch} &0.94\% & 3.825&0.024\% &51.40	&--   & --\\
		TNLS~\cite{LiPermutationSearch} &  1.11\% &0.673&	0.038\% &59.83	&--   & --\\
		TNALE~\cite{li2023alternating} &  1.65\% &0.201&0.047\% &19.96&--&--\\
		SVDinsTN~  & 1.13\% &\textbf{0.002}	& \underline{0.016\%}	&	\textbf{0.017}	& \underline{0.0007\%}	& \textbf{0.608}\\	
		
		\bottomrule
	\end{tabular}	\vspace{-0.2cm}
\end{table}

\begin{table*}[!t]
	\small
	\setlength{\tabcolsep}{2.25pt}
	\renewcommand\arraystretch{1}
	\centering
	\caption{Comparison of MPSNR ($\uparrow$) and run time (in seconds, $\downarrow$) of different TC methods on color videos.}\label{CVtab2}
	\begin{tabular}{ccccccccccccccc}
		\toprule
		\multirow{2}{*}{Video}         & \multicolumn{2}{c}{\,\,FBCP~\cite{BayesianCP}\,\,}    & \multicolumn{2}{c}{\,\,TMac~\cite{Xu2015Tmac}\,\,}   & \multicolumn{2}{c}{TMacTT~\cite{TIPTT}}  & \multicolumn{2}{c}{\,TRLRF~\cite{yuan2019tensor}\,} &  \multicolumn{2}{c}{TW~\cite{NEURIPS2022_acbfe708}} &  \multicolumn{2}{c}{TNLS~\cite{LiPermutationSearch}}& \multicolumn{2}{c}{SVDinsTN} \\
		
		\cmidrule(rl){2-3}\cmidrule(rl){4-5} \cmidrule(rl){6-7}\cmidrule(rl){8-9}\cmidrule(rl){10-11}\cmidrule(rl){12-13}\cmidrule(rl){14-15}
		&  MPSNR  & Time &  MPSNR & Time & MPSNR  & Time  & MPSNR  & Time  & MPSNR  & Time & MPSNR  & Time & MPSNR  & Time \\
		
		\midrule
		
		\emph{Bunny}  &28.402  & 1731.2  & 28.211  & 1203.5  &29.523 & \underline{453.76} & 29.163 & 486.76  &30.729&1497.4&28.787 &99438 & \textbf{32.401}& 691.33 \\
		\emph{News}   &28.234& 1720.4& 27.882 & \underline{340.46} & 28.714 & 535.97 & 28.857  & 978.12 &30.027&1426.3&29.761 &37675&\textbf{31.643} & 932.42\\
		
		\emph{Salesman} &29.077& 1783.2 & 28.469 &\underline{353.63} & 29.534  &656.45& 28.288  &689.35&30.621&1148.7&30.685&76053&\textbf{31.684}& 769.54\\

		\emph{Silent} &30.126 &1453.9 & 30.599 & \underline{316.21} & 30.647&1305.6  & 31.081& 453.24 &31.731& 1232.0&28.830&98502&\textbf{32.706} & {532.31} \\
		
		\bottomrule
	\end{tabular}
\end{table*}

\subsection{Experiments for Validating Claim C}\label{ExperC}

\begin{figure*}[!t]
	\small
	\setlength{\tabcolsep}{1.5pt}
	\renewcommand\arraystretch{1}
	\centering
	\begin{tabular}{cccccccc}
		FBCP~\cite{BayesianCP}&TMac~\cite{Xu2015Tmac}&TMacTT~\cite{TIPTT}&TRLRF~\cite{yuan2019tensor} &TW~\cite{NEURIPS2022_acbfe708}  & TNLS~\cite{LiPermutationSearch} & SVDinsTN  & Ground truth\\[0.025cm]
		\includegraphics[width=0.118\textwidth]{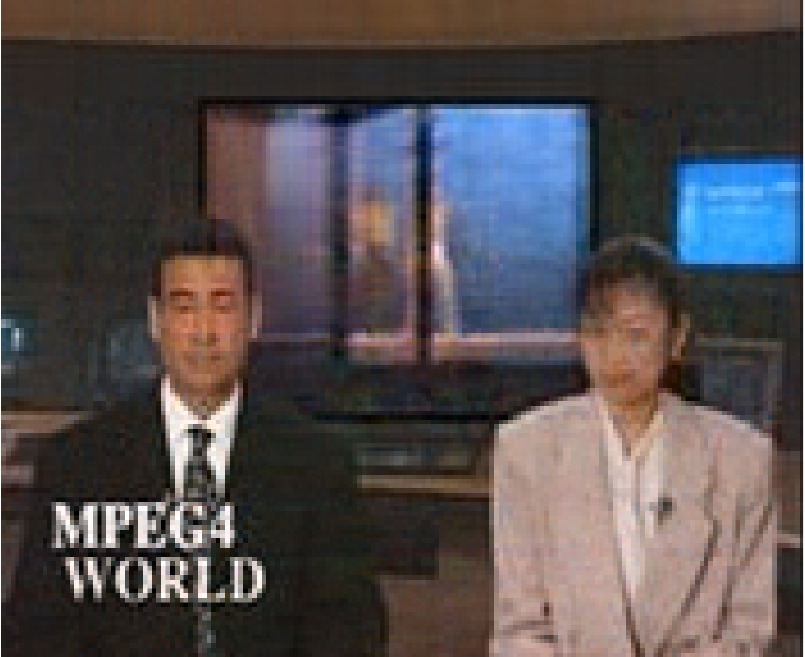}&
		\includegraphics[width=0.118\textwidth]{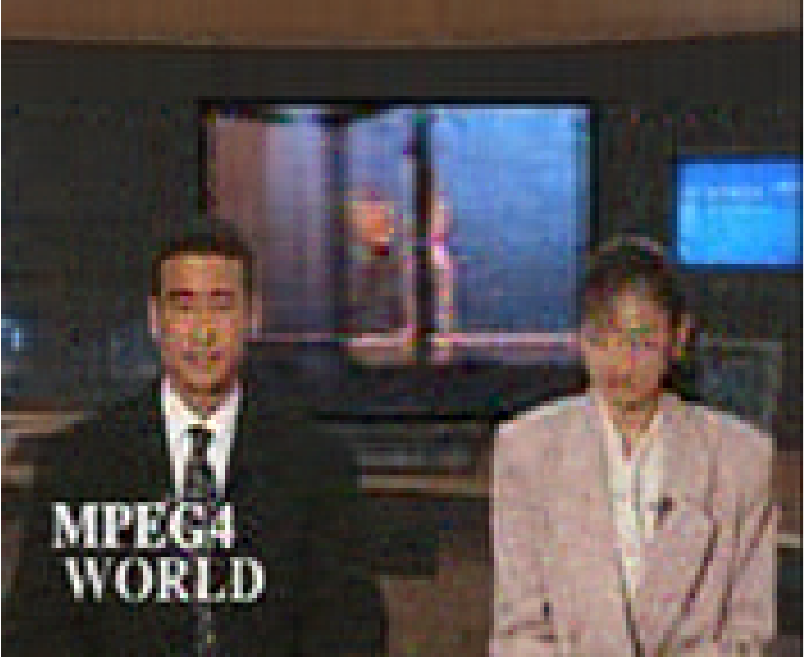}&
		\includegraphics[width=0.118\textwidth]{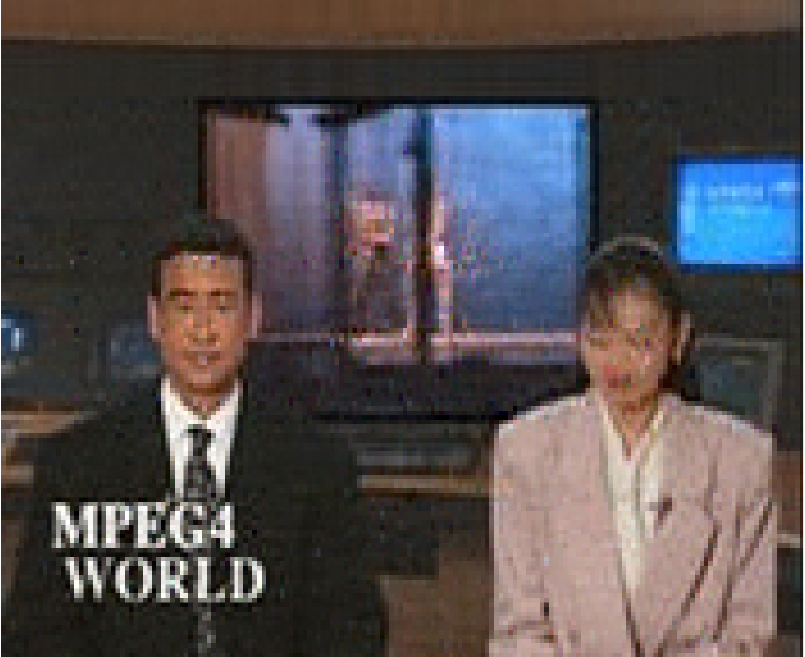}&
		\includegraphics[width=0.118\textwidth]{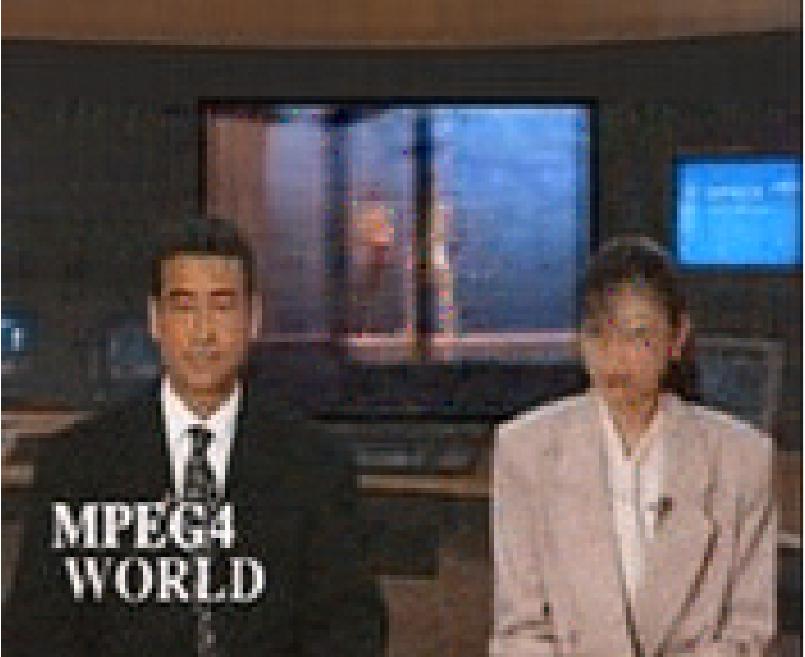}&
		\includegraphics[width=0.118\textwidth]{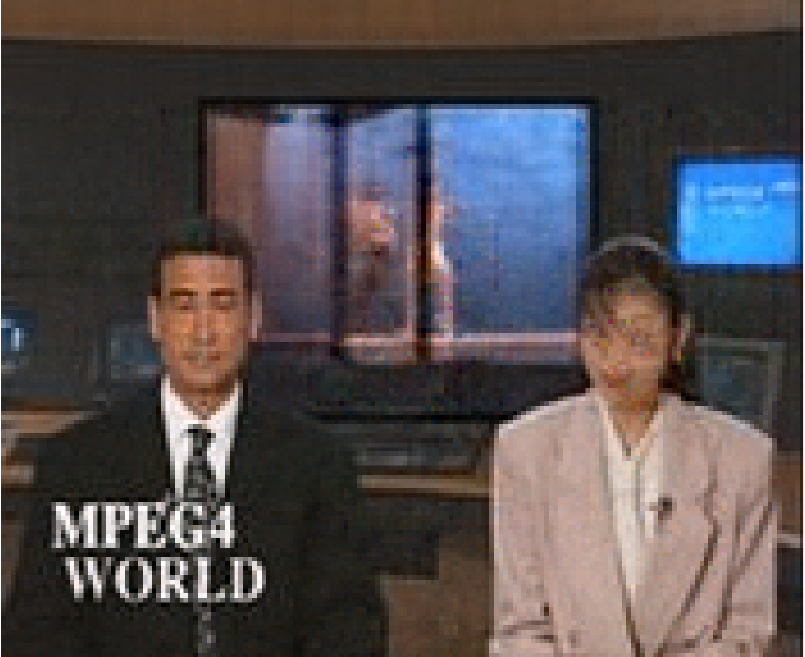}&
		\includegraphics[width=0.118\textwidth]{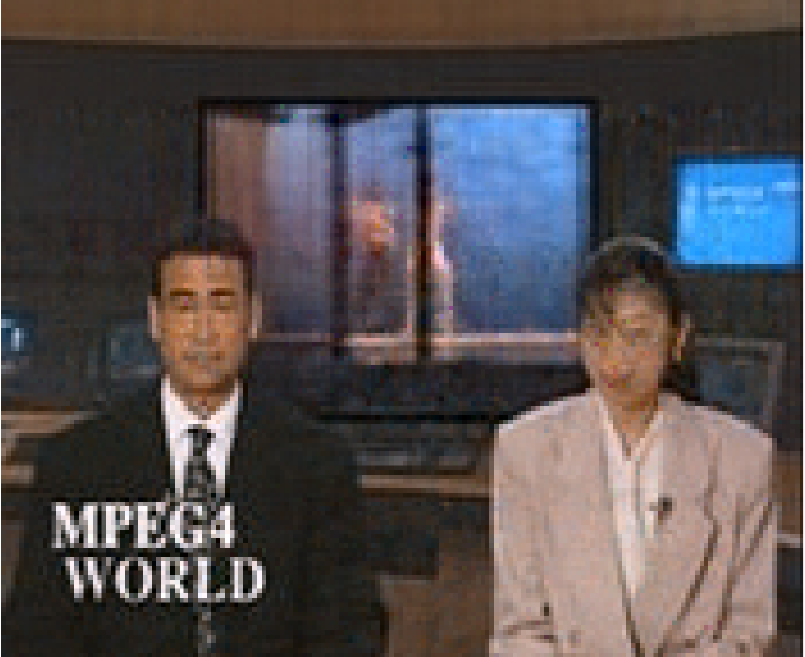}&
		\includegraphics[width=0.118\textwidth]{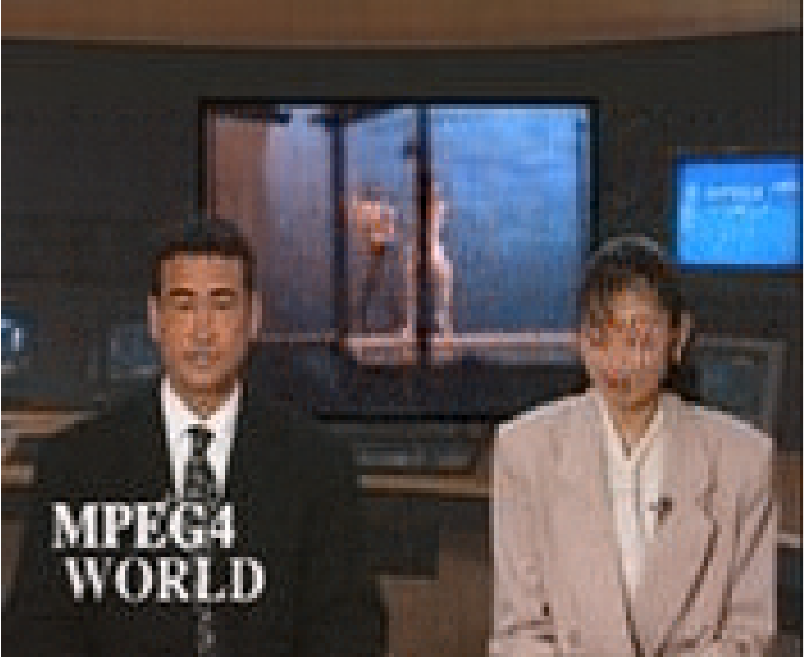}&
		\includegraphics[width=0.118\textwidth]{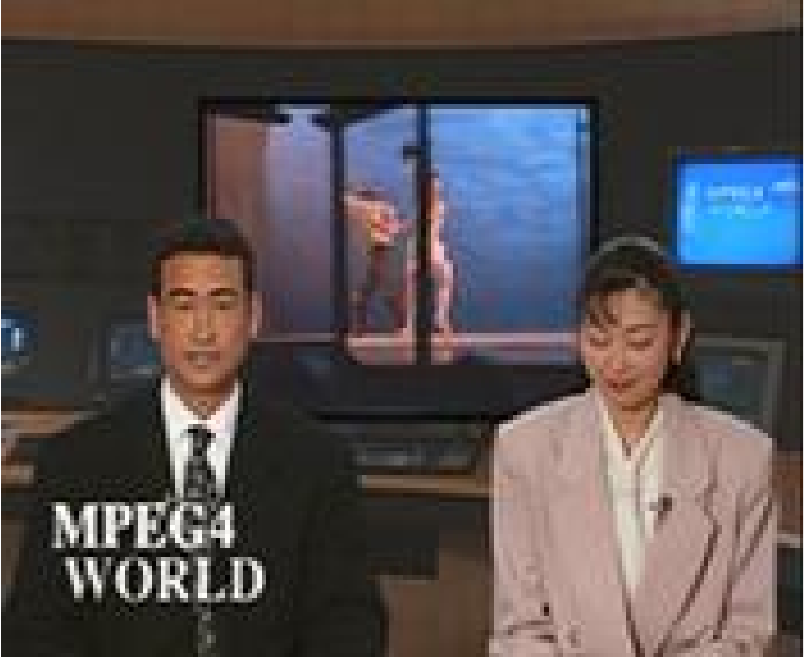}\\[0.05cm]
		\includegraphics[width=0.118\textwidth]{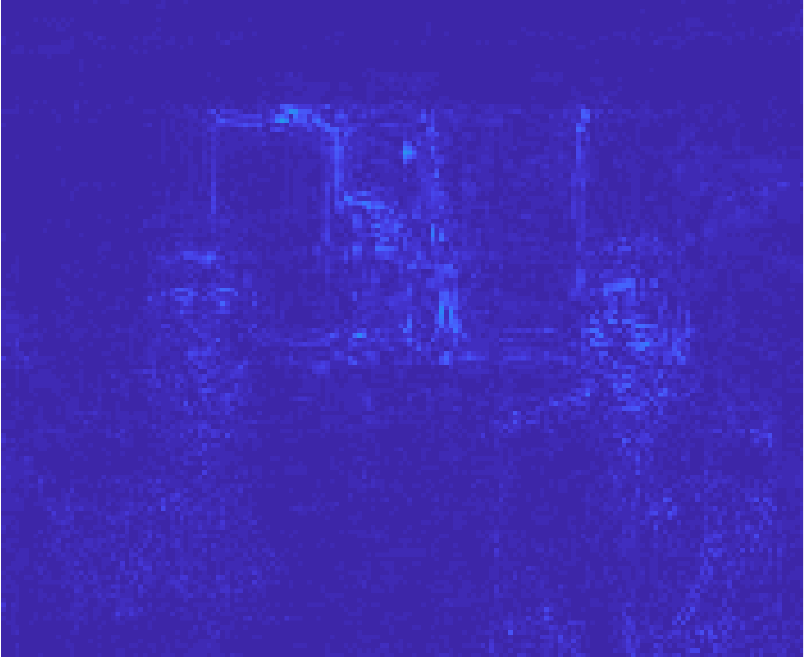}&
		\includegraphics[width=0.118\textwidth]{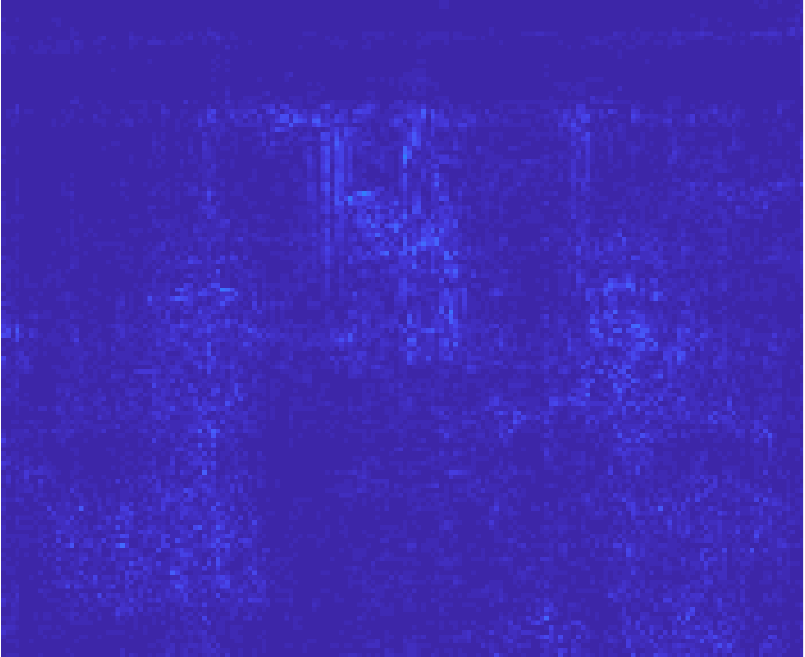}&
		\includegraphics[width=0.118\textwidth]{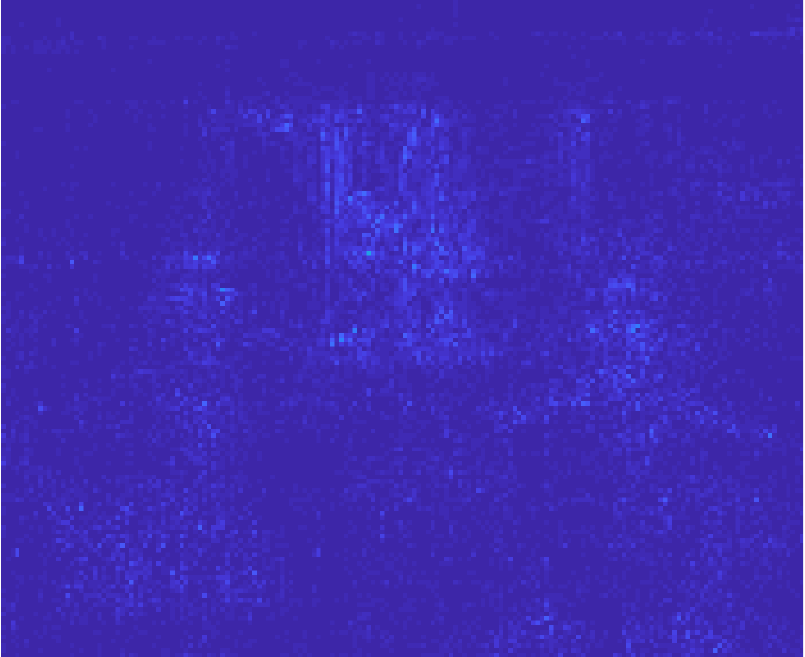}&
		\includegraphics[width=0.118\textwidth]{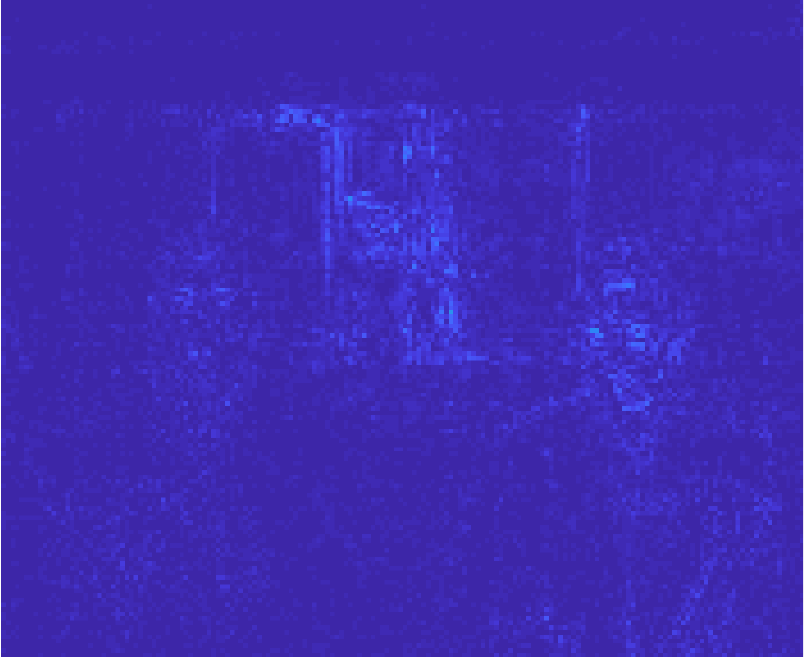}&
		\includegraphics[width=0.118\textwidth]{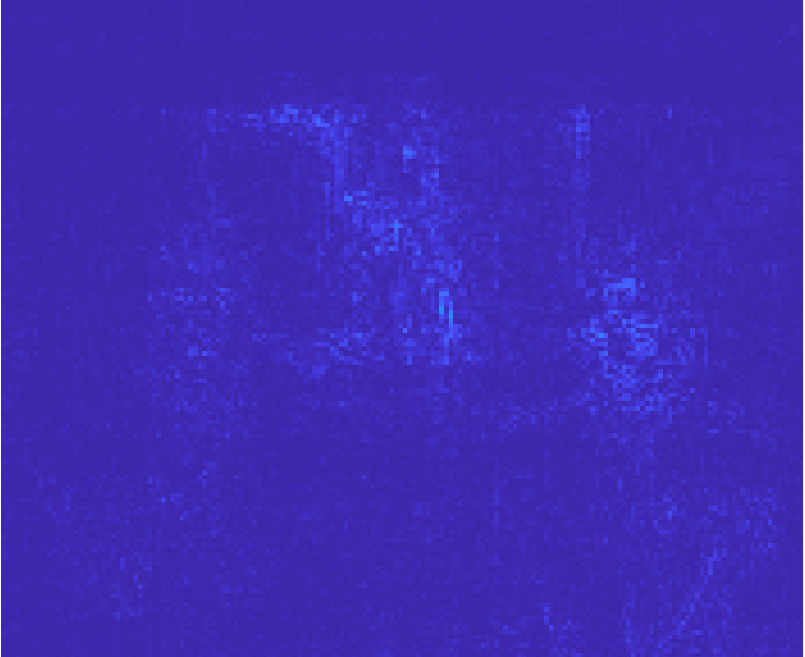}&	
		\includegraphics[width=0.118\textwidth]{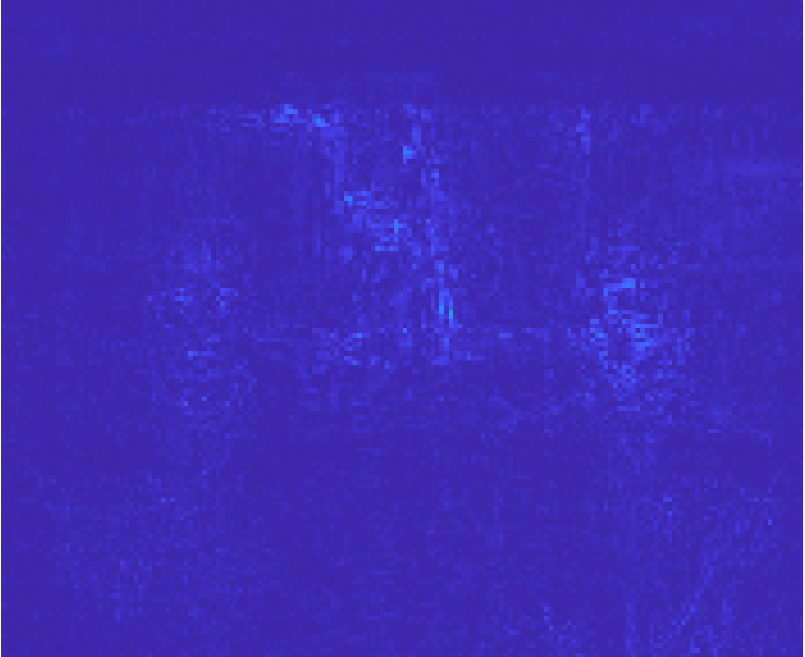}&	
		\includegraphics[width=0.118\textwidth]{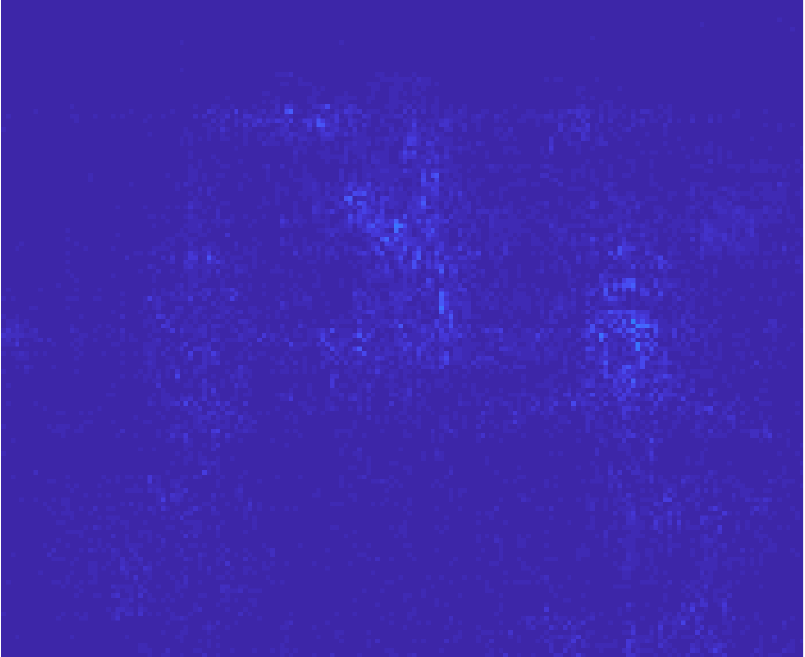}&
		\includegraphics[width=0.118\textwidth]{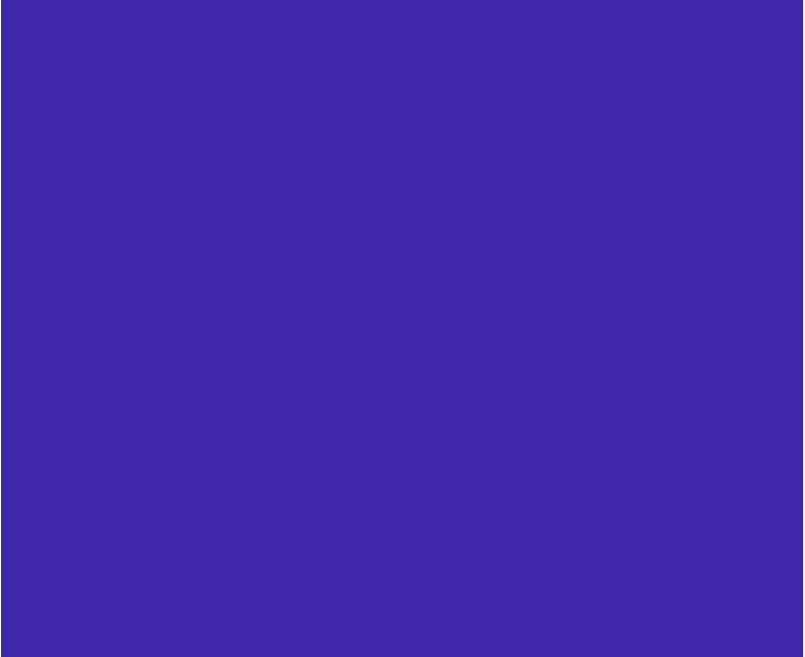}\\[0.05cm]
		\multicolumn{8}{c}{\includegraphics[width=0.95\textwidth]{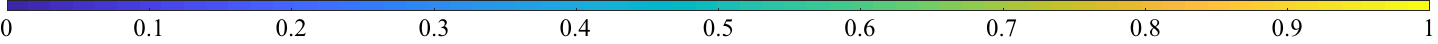}}
	\end{tabular}
	\caption{Reconstructed images and residual images obtained by different methods on the 25th frame of \emph{News}. Here the residual image is the average absolute difference between the reconstructed image and the ground truth over R, G, and B channels.}\label{CVfig}	\vspace{-0.15cm}
\end{figure*}

We conduct experiments to validate Claim C. We employ the proposed SVDinsTN to a fundamental application, i.e., tensor completion (TC), and compare it with the state-of-the-art tensor decomposition-based TC methods. Given an incomplete observation tensor $\mathcal{F}\in \mathbb{R}^{I_1\times I_2\times\cdots\times I_N}$ of $\mathcal{X}\in \mathbb{R}^{I_1\times I_2\times\cdots\times I_N}$, the proposed TC method first updates $\mathcal{G}$ and $\mathbf{S}$ by Algorithm \ref{AlgforSTNm}, and then updates the target tensor $\mathcal{X}$ as follows: $\mathcal{X}=\mathcal{P}_{\Omega^c}((\text{STN}(\mathcal{G},\mathbf{S})+\rho\hat{\mathcal{X}})/(1+\rho))+\mathcal{P}_{\Omega}(\mathcal{F})$, where $\Omega$ is the index set of the known elements, $\mathcal{P}_{\Omega}(\mathcal{X})$ is a projection operator that projects the elements in $\Omega$ to themselves and all others to zeros, $\hat{\mathcal{X}}$ is the result at the previous iteration, and the initial ${\mathcal{X}}$ is $\mathcal{F}$.

\textbf{Experiment setting.} We test four color videos\footnote{\url{http://trace.eas.asu.edu/yuv/}}, named \textit{Bunny}, \textit{News}, \textit{Salesman}, and \textit{Silent}, which are fourth-order tensors of size $144\times 176\times 3\times 50$ (spatial height $\!\times\!$ spatial width $\!\times\!$ color channel $\!\times\!$ frame). We employ six methods for comparison, named FBCP~\cite{BayesianCP}, TMac~\cite{Xu2015Tmac}, TMacTT~\cite{TIPTT}, TRLRF~\cite{yuan2019tensor}, TW~\cite{NEURIPS2022_acbfe708}, and TNLS\footnote{TNLS excels in the compression task; therefore, we use it as a representative TN-SS method for comparison.}~\cite{LiPermutationSearch}, respectively. We set the missing ratio (MR) to $90\%$, which is defined as the ratio of the number of missing elements to the total number of elements. We evaluate the reconstructed quality by \textit{the mean peak signal-to-noise ratio (MPSNR)} computed across all frames. In all tests, the parameter $\gamma$ is set to $0.0003$.% The parameter setting of the baseline methods is provided in the \textit{Supplementary Material}. 

\textbf{Result analysis.} Table \ref{CVtab2} reports \textit{MPSNR} and \textit{run time} obtained by different TC methods. As observed, the proposed SVDinsTN consistently achieves the highest \textit{MPSNR} values among all utilized TC methods across all test color videos. In Figure \ref{CVfig}, we present the reconstructed images and their corresponding residual images at the 25th frame of \textit{News}. We observe that the proposed SVDinsTN outperforms the baseline methods in terms of visual quality, particularly with respect to background cleanliness and local details (\eg ``dancer'') recovery.

\begin{figure}[!t]
	\setlength{\tabcolsep}{1.2pt}
	\renewcommand\arraystretch{1}
	\centering
	\begin{tabular}{cc}
		\includegraphics[width=0.235\textwidth]{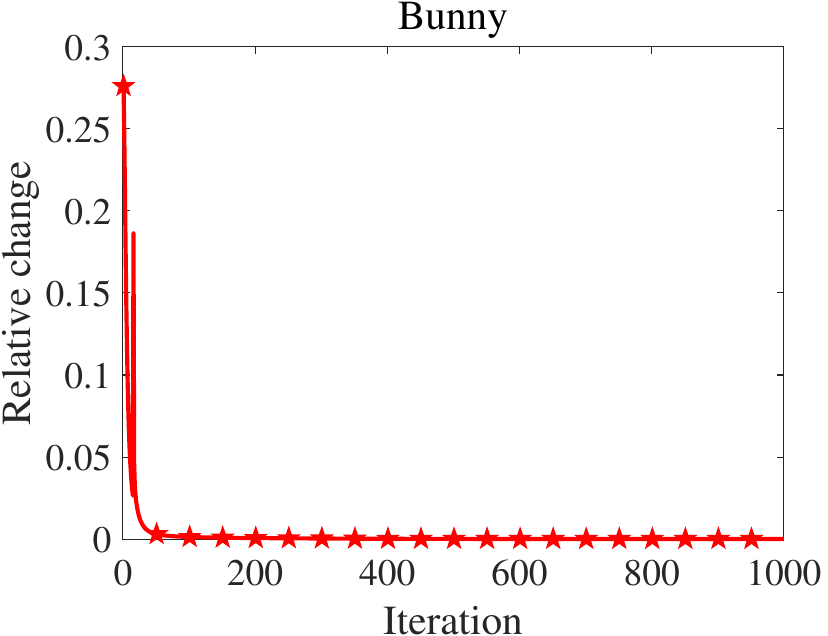}&
		\includegraphics[width=0.235\textwidth]{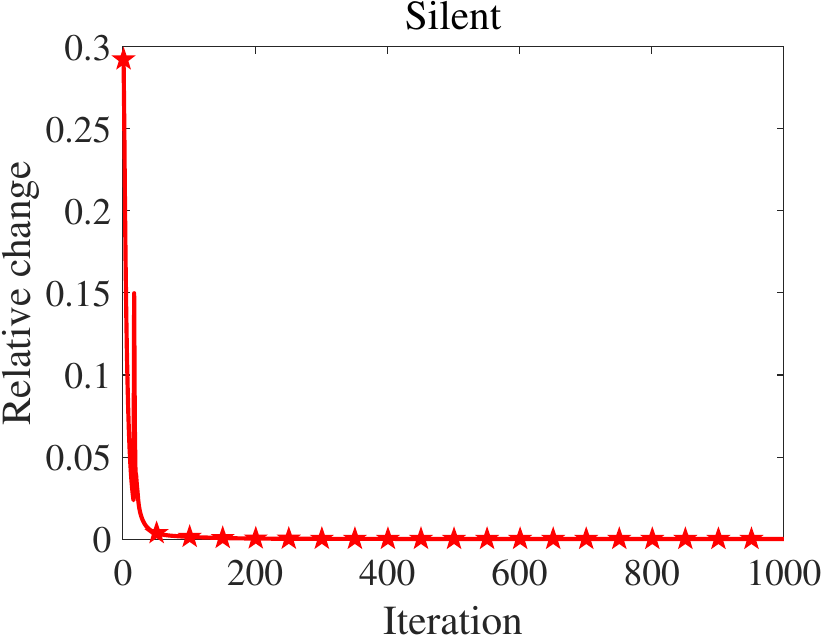}\\
	\end{tabular}
	\caption{Relative change curves with respect to the iteration number on test color videos \textit{Bunny} and \textit{Silent}. Here the relative change is defined as $\|\mathcal{X}-\hat{\mathcal{X}}\|_F/\|\hat{\mathcal{X}}\|_F$, and $\mathcal{X}$ and $\hat{\mathcal{X}}$ are the results of the current iteration and its previous iteration.}\label{CVcon}	\vspace{-0.3cm}
\end{figure}

\textbf{Numerical convergence.} In Theorem~\ref{Conver}, we provide a \textit{theoretical} convergence guarantee for the proposed method. Here, we select color videos \textit{Bunny} and \textit{Silent} as examples to \textit{numerically} verify the convergence. Figure~\ref{CVcon} presents the relative change in the reconstructed color videos at each iteration compared to their respective previous iterations. We observe that the values of the relative change achieved by the proposed method decrease and gradually tend to zero as the number of iterations increases. This justifies the numerical convergence of the proposed method.

\section{Conclusion} \label{Conclusion}

We propose a novel TN paradigm, called SVDinsTN, which enables us to solve the challenging TN-SS problem from a regularized modeling perspective. This perspective renders our model highly amenable to easy solutions, allowing us to leverage well-established optimization algorithms to solve the regularized model. As a result, the proposed method achieves about $100\!\sim{}\!1000$ times acceleration compared to the state-of-the-art TN-SS methods with a comparable representation ability. Besides, SVDinsTN demonstrates its effectiveness as a valuable tool in practical applications.  

\textbf{Limitations.} In existing research on TN-SS, two challenging issues remain open. One is the computationally consuming issue, and the other is the theoretical guarantee of the optimal TN structure. SVDinsTN addresses the computationally consuming issue. But the theoretical guarantee of the optimal TN structure is still an open problem. Solving this issue will be the direction of our future work.

\section*{Acknowledgements} 

We would like to express our gratitude to Prof. Guillaume Rabusseau for his valuable assistance in correcting the experimental results of ``TNGreedy''.

{
    %\small
    \bibliographystyle{ieeenat_fullname}

}


\end{document}